\documentclass[10pt,onecolumn,letterpaper]{article}
\usepackage{multicol}
\usepackage{iccv}
\usepackage{times}
\usepackage{epsfig}
\usepackage{graphicx}
\usepackage{amsmath}
\usepackage{amssymb}
\usepackage{comment}
\usepackage{array}
\usepackage{enumitem}
\usepackage{lipsum}

\usepackage{cuted}
\usepackage{capt-of}
\usepackage{multirow}
\usepackage{comment}
\usepackage{array}
\usepackage{enumitem}
\usepackage{algorithm2e}
\usepackage{pdflscape}
\usepackage{color, colortbl}
\usepackage[dvipsnames]{xcolor}
\usepackage{ragged2e}
\usepackage[pagebackref=true,breaklinks=true,letterpaper=true,colorlinks,bookmarks=false]{hyperref}
\newcommand{\ImWidth}{42mm}
\newcommand{\RowHeight}{5pt}
\newcommand{\RotateAngle}{90}
\newcommand\fontsizesmallemail{\fontsize{8.7pt}{8.7pt}\selectfont}
\usepackage{longtable}
\usepackage{adjustbox}
\usepackage{caption}
\usepackage{subcaption}
\DeclareCaptionFont{customfontsize}{\fontsize{8.97}{11}\selectfont}
\usepackage{multicol}
\iccvfinalcopy 

\def\iccvPaperID{****} 

\begin{document}
\definecolor{commentcolor}{RGB}{0,0,255}
\newcommand{\PyComment}[1]{\ttfamily\textcolor{commentcolor}{\# #1}}  
\newcommand{\PyCode}[1]{\ttfamily\textcolor{black}{#1}} 

\title{SATIN: A Multi-Task Metadataset for Classifying Satellite Imagery using Vision-Language Models}

\author{Jonathan Roberts\\
University of Cambridge\\
{\tt\fontsizesmallemail jdr53@cam.ac.uk}
\and
\hspace{-5mm}
Kai Han\\
The University of Hong Kong\\
{\tt\fontsizesmallemail kaihanx@hku.hk}
\and
\hspace{-5mm}
Samuel Albanie\\
University of Cambridge\\
{\tt\fontsizesmallemail samuel.albanie.academic@gmail.com}
}

\maketitle
\ificcvfinal\thispagestyle{empty}\fi
\vspace{-0.6cm}
\begin{figure*}[h!]
\centering
  \includegraphics[width=0.97\textwidth]{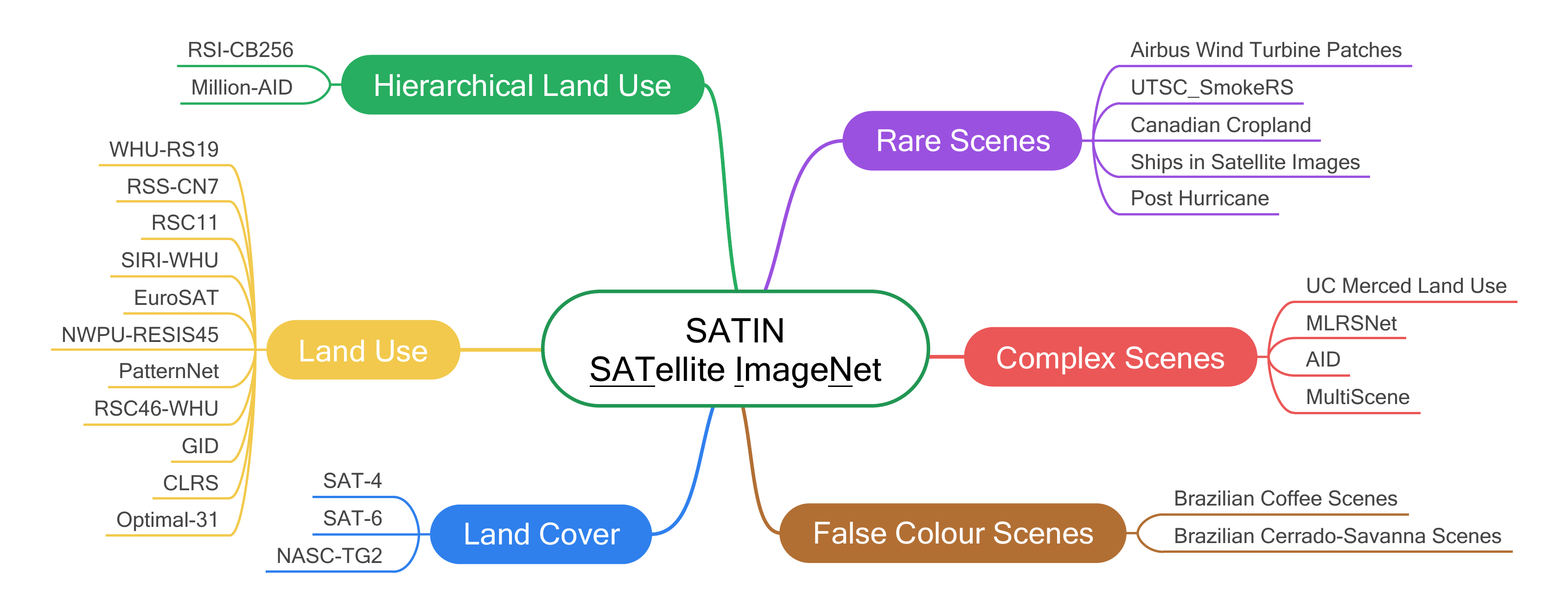}
  \vspace{-2mm}
  \captionof{figure}{\textbf{The SATIN taxonomy}. We propose the SATIN benchmark containing 27 constituent datasets spanning 6 distinct tasks. 
  The imagery is globally distributed, comprised of resolutions spanning 5 orders of magnitude 
   and over 250 distinct class labels.
  }
  \label{fig:teaser}
\end{figure*}

\begin{multicols}{2}
\begin{abstract}
Interpreting remote sensing imagery enables numerous downstream applications ranging from land-use planning to deforestation monitoring. Robustly classifying this data is challenging due to the Earth's geographic diversity. While many distinct satellite and aerial image classification datasets exist, there is yet to be a benchmark curated that suitably covers this diversity.
In this work, we introduce \underline{SAT}ellite \underline{I}mage\underline{N}et (SATIN), a metadataset curated from 27 existing remotely sensed datasets, and comprehensively evaluate the zero-shot transfer classification capabilities of a broad range of vision-language (VL) models on SATIN. We find SATIN to be a challenging benchmark---the strongest method we evaluate achieves a classification accuracy of 52.0\%. We provide a public leaderboard\footnote{\url{https://satinbenchmark.github.io}} to guide and track the progress of VL models in this important domain. 

\vspace{-1mm}
\end{abstract}
\section{Introduction}

Remote sensing (RS) of the Earth's surface offers the potential for efficient provision of continuous sensor data at a range of temporal and spatial scales~\cite{congalton2014global}. The ability to interpret this imagery yields numerous benefits and downstream applications spanning land-use planning, natural resource management and food security \cite{saah_landcover_examples, fritz2010comparison} that can inform public policy~\cite{cihlar2000land} and enable management of environmental risk~\cite{KERR2003299}. Moreover, the (near) continuous monitoring enables detection of changes such as deforestation \cite{grinand2013estimating}, forest degradation \cite{mitchell2017current}, forecasting sea ice \cite{makynen2020satellite} and modelling urban sprawl \cite{boori2015monitoring}.

A core component of these applications is \mbox{\textit{classification}} -- the ability to label an image, pixel or sub-pixel according to a set of categories. This task is challenging in the RS domain due to a number of reasons: 
(1) \textbf{Image diversity}---Satellite imagery, the predominant RS data source, is captured and released at a variety of fields of view and resolutions \cite{kim2020new, wagner2018geographic, earthnets4eo, long2021creating, rogan2004remote}. Furthermore, the global coverage of satellite imagery reflects the Earth's wide range of geographic diversity. 
(2) \textbf{Label hierarchies}---Land cover and land use \cite{comber2008land} can be labeled according to different levels of abstraction: for example, a paddy field can be classed as \textit{natural} $\rightarrow$ \textit{agricultural land} $\rightarrow$ \textit{arable land} $\rightarrow$ \textit{paddy field} (e.g., \cite{long2021creating}). 
(3) \textbf{Scene complexity}---The large scale of satellite imagery means that scenes can be rich in detail and include multiple different objects and land use classes. 
These challenges, coupled with the lack of a streamlined benchmark for RS has precluded development in this domain. Thus, there is a strong need for a satellite imagery \textit{metadataset} -- a benchmark that consists of many existing datasets grouped into key tasks -- to guide the development and progress of interpreting RS imagery. 

To date, the curation of a satellite imagery metadataset has been hindered by two key challenges: 
(i) \textbf{Models}---State-of-the-art supervised methods, such as convolutional neural networks, struggle to robustly classify satellite imagery across different domains and sets of categories, and to ingest imagery in disparate formats. Copious amounts of both compute and annotated data are required to overcome these limitations. 
(ii) \textbf{Data}---Datasets can be difficult to access and have inconsistent descriptions (e.g., lack key information on image resolution and geographic area). Additionally, it can be challenging to store and evaluate on multiple datasets containing different file structures, formats and other idiosyncrasies. 

Due to the following recent advancements, curating an RS classification metadataset is particularly timely: 
(1) \textbf{VL pretraining}---The emergence of seminal open-vocabulary models such as CLIP \cite{radford2021learning} and ALIGN \cite{jia2021scaling} enables 
zero-shot transfer: classification on unseen data distributions using natural language. This transfer paradigm obviates the need for fixed category labels, enabling a single VL model to be evaluated across a suite of tasks and datasets.
(2) \textbf{Platforms \& APIs}---The introduction of data and model hosting platforms and APIs (such as HF datasets~\cite{lhoest-etal-2021-datasets}, TF Datasets~\cite{TFDS} and HF Transformers~\cite{wolf-etal-2020-transformers}) provide simple and inexpensive pipelines for hosting, downloading and evaluating models and datasets. 
(3) \textbf{RS data}---As the availability, global coverage (every few days \cite{dubovik}) and resolution of RS data continue to advance, there is a 
need for algorithms that can interpret this data accurately and robustly.

In this work, we introduce SATIN (\underline{SAT}ellite \underline{I}mage\underline{N}et), a metadataset curated from 27 existing datasets spanning a variety of tasks, resolutions, fields of view, and geographic areas. We leverage platforms such as HF datasets to create a streamlined benchmark that can be used and evaluated smoothly and seamlessly with minimal friction.
Furthermore, we evaluate the performance of a selection of open-source VL baselines on SATIN and find that SATIN proves to be a challenging zero-shot benchmark, with even the highest capacity models scoring just above 50\% accuracy. To encourage and keep track of progress in the RS domain, we make a public leaderboard available.

In summary, our contributions are four-fold:
\begin{enumerate}[label={(\roman*)}]
    \vspace{-0.2cm}\item We curate a challenging multi-task metadataset of remote sensing imagery;
    \vspace{-0.2cm}\item We collate key information on a variety of existing datasets and release them;
    \vspace{-0.2cm}\item We benchmark a broad profile of VL models on our metadataset, showcasing that even the best ones find it challenging;  
    \vspace{-0.2cm}\item We provide a public leaderboard for the benchmark.
\end{enumerate}

\section{Related Work}
Our work is related to several themes in the literature including \textit{remote sensing classification datasets}, \textit{metadatasets}, and \textit{VL model evaluation}.

\noindent\textbf{Remote sensing classification datasets}.
In recent years, numerous RS classification datasets have been released \cite{earthnets4eo, long2021creating}. Popular works include UC Merced Land Use \cite{yang2010bag}, EuroSAT \cite{helber2019eurosat} and WHU-RS19 \cite{Xia2010WHURS19, Dai2011WHURS19}, which provide a useful benchmark for machine learning models, though are relatively small and limited to a single resolution, field of view size, and geographic region. Recently, larger-scale datasets have been released with globally distributed imagery (e.g., \cite{christie2018functional, long2021creating, qi2020mlrsnet, cornebise2022open, zhu2019so2sat, sumbul2019bigearthnet}) and multiple resolutions and image sizes. The SATLAS dataset \cite{bastani2022satlas} improves upon these with increased scale and number of labels, and by including different modalities, while it is still limited in variety of resolutions, image sizes and includes only a narrow classification task. Million-AID \cite{long2021creating} provides an extensive land use benchmark that promotes image diversity, richness and scalability. Our SATIN metadataset agglomerates 27 of these relatively disjointed datasets, providing a comprehensive benchmark for the field of RS image classification spanning a broad range of resolutions, fields of view and tasks, and comprehensively reflects the breadth of the challenge of RS image classification.

\noindent\textbf{Metadatasets}.
In recent years, metadatasets have emerged as useful and impactful in various fields. In natural language processing, metadatasets such as GLUE \cite{wang2018glue}, SuperGLUE \cite{wang2019superglue} and EleutherAI LM Harness \cite{eval2021harness} have been used to evaluate the performance of models on a range of existing datasets, allowing for a more comprehensive understanding of their strengths and limitations. Furthermore, works such as HELM \cite{liang2022holistic} and BIG-bench \cite{srivastava2022beyond} were introduced as \textit{living benchmarks} that are updated as new scenarios, tasks, metrics and models are added. In computer vision, metadatasets such as VTAB \cite{zhai2019large}, META-DATASET \cite{triantafillou2019meta}, ELEVATER \cite{li2022elevater} and NEVIS'22 \cite{bornschein2022nevis} similarly provide an evaluation for the progress of classification models. However, META-DATASET does not include any RS datasets, and ELEVATER and VTAB only include two -- both are small-scale datasets that do not adequately reflect the diversity of the domain; NEVIS'22 also only contains two very small niche RS datasets.
SATIN, on the other hand, includes 27 remote sensing datasets and is the first metadataset curated for remote sensing. One notable concurrent work is AiTLAS \cite{DIMITROVSKI202318}, which provides a benchmark against remote sensing classification datasets.
Our work makes important distinctions in that (i) we curate a metadataset consisting of different tasks;
(ii) the constituent datasets are collected and redistributed;
(iii) a wide array of VL models are evaluated and a public leaderboard is further provided for tracking future progress in the RS domain.

\noindent\textbf{Evaluation of VL models}. 
The evaluation of VL classification models on remote sensing datasets has thus far been limited to the inclusion of a few remote sensing datasets as part of wider profiles of image classification datasets (most of which contain natural images), e.g., \cite{radford2021learning, jia2021scaling, li2022elevater, zhai2019large, yao2021filip, mu2022slip}. Prior to our work, there has not been a broad profile evaluation of the capacity of VL models to interpret and classify remote sensing imagery.
\section{SATIN Overview}

The SATIN metadataset has been constructed to provide an evaluation of the capabilities of modern machine learning models on a broad range of RS image classification tasks. We intend for this benchmark to (i) primarily be used to evaluate VL models in a zero- or low-shot classification setting, and (ii) be a \textit{living benchmark} with incremental releases offering expansions and improvements. 

In the following, we detail how we build SATIN (Sec.~\ref{satin:design-process}) from multiple datasets~(Sec.~\ref{satin:datasets}), what tasks are included in the benchmark (Sec.~\ref{satin:tasks}), and considered evaluation metrics (Sec.~\ref{satin:metrics}).

\subsection{Design Process}\label{satin:design-process}
SATIN has been curated from a wide set of preexisting datasets to maximally increase the variety of resolutions, field of view sizes, geographic locations, label categories and tasks. In our dataset selection, we have optimised for this diversity, and aimed to include as many datasets as possible, subject to the following constraints: 
\begin{itemize}
    \vspace{-0.1cm}
    \item \textbf{Licences}. We do not include datasets with research usage licencing restrictions (details given in App. \ref{appendix:licencing_details}). 
    \vspace{-0.1cm}
    \item \textbf{Format}. We focus on 3-channel RGB imagery, except for the \textit{False Colour Scenes} task datasets, which include Near-Infrared, Red, Green. Some of the selected datasets include RGB and other bands, in which cases we only include the RGB imagery. 
    \vspace{-0.1cm}
    \item \textbf{Platform}. Only imagery derived from satellite or airplane platforms is used, i.e., drone imagery datasets are not included. Multi-view datasets with aerial and corresponding `street-level' imagery are excluded.
    \vspace{-0.2cm}
    \item \textbf{Quality}. Based on manual visual inspection, datasets containing poor-quality imagery and/or annotations are rejected.
    \vspace{-0.2cm}
    \item \textbf{Classification}. Only datasets with image-level labels are included.
    \vspace{-0.2cm}
    \item \textbf{Size}. We avoid large datasets ($\gg$100,000 images) as: (1) We focus on the low-shot domain, in which lower volumes are sufficient. (2) Democratisation -- more of the community can benefit from a smaller-scale benchmark. (3) We increase image diversity as much as possible at a given data storage budget by having more smaller datasets rather than fewer larger ones.
\end{itemize}

\subsection{Datasets}\label{satin:datasets}
Table \ref{Table:datasets_overview} gives details of the 27 datasets included in the SATIN metadataset. Holistically, SATIN aims to reflect the diversity in RS data by including imagery that is globally distributed, comprised of resolutions spanning from 0.06 m to 1000 m, multiple fields of view sizes, and over 250 distinct category labels. While we include the majority of the datasets used in SATIN in their entirety (regardless of predefined splits), for some larger datasets (denoted in the table with an *) we employ subsets defined by the dataset creators (details of the subsets we use are given in App. \ref{appendix:subsets}).

\subsection{Tasks}\label{satin:tasks}
\begin{figure*}
\scalebox{0.9}{
\begin{tabular}{ccp{0.01cm}ccp{0.01cm}cc}
  \includegraphics[width=32mm]{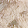} &  \includegraphics[width=32mm]{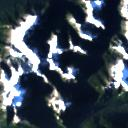} 
  & & 
  \includegraphics[width=32mm]{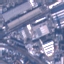} & \includegraphics[width=32mm]{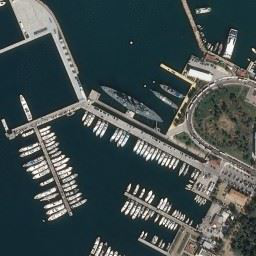} 
  & & \multicolumn{2}{c}{\includegraphics[width=32mm]{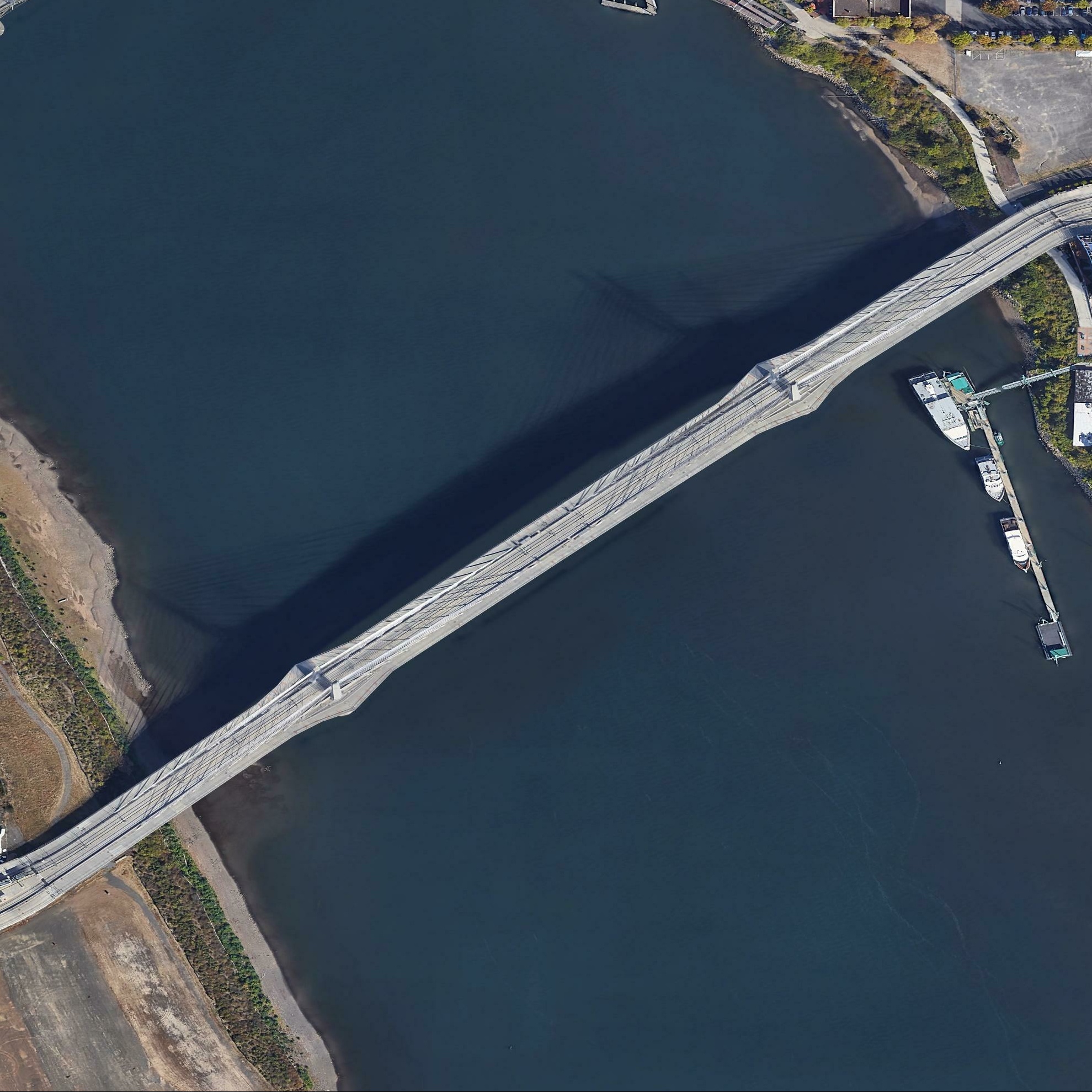}}
  \\ 
barren & snowberg 
& & industrial & port 
& & \multicolumn{2}{c}{\begin{tabular}[c]{@{}c@{}} \textit{transportation land $\rightarrow$}\\ \textit{highway area $\rightarrow$ bridge}\end{tabular}} \\[6pt]
\multicolumn{2}{c}{Task 1. \textbf{Land Cover}} 
& & \multicolumn{2}{c}{Task 2. \textbf{Land Use}} 
& & \multicolumn{2}{c}{Task 3. \textbf{Hier. Land Use}} \\[6pt]
  \includegraphics[width=32mm]{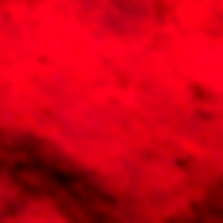} & \includegraphics[width=32mm]{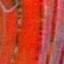}
  & & \includegraphics[width=32mm]{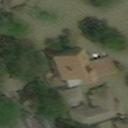} & \includegraphics[width=32mm]{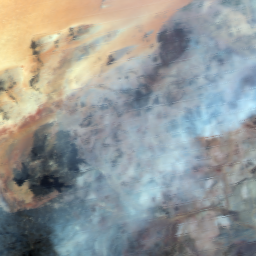} 
  & & 
  \multicolumn{2}{c}{
  \includegraphics[width=32mm]{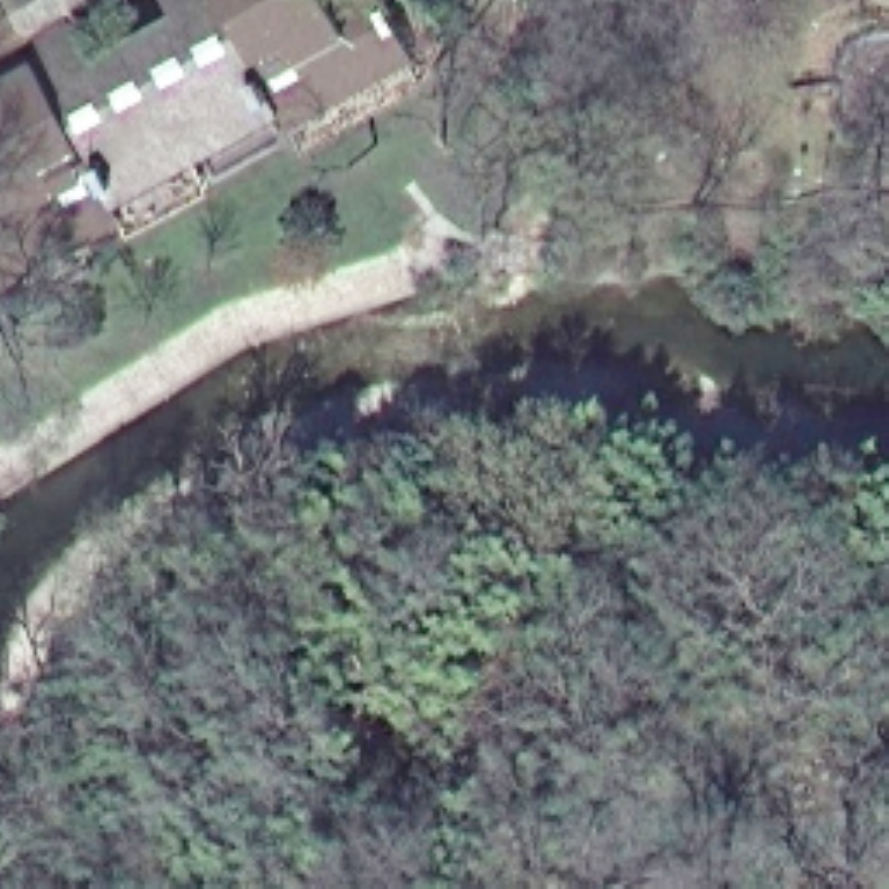}}
  \\
\textit{shrubby vegetation} & \textit{coffee} 
& & \textit{flooded/damaged} & \textit{dust} 
& & \multicolumn{2}{c}{\begin{tabular}[c]{@{}c@{}} \textit{bare soil, buildings, grass}\\ \textit{sand, trees, water}\end{tabular}}
\\[6pt]
\multicolumn{2}{c}{Task 6. \textbf{False-Colour Scenes}}
& & \multicolumn{2}{c}{Task 5. \textbf{Rare Scenes}} 
& & \multicolumn{2}{c}{Task 4. \textbf{Complex Scenes}} \\[6pt]
\end{tabular}
}
\vspace{-0.5cm}
\caption{\textbf{Example imagery and ground truth labels for each SATIN task.} This subset includes imagery from 10 of the 27 constituent datasets and shows the breadth of diversity in SATIN across different imagery types, resolutions, fields of view and geographic areas.
}

\label{Table:task_overview}
\end{figure*}

We group the selected datasets into 6 tasks, see Table \ref{Table:task_overview} for examples of each task.\footnote{Owing to the ambiguity and overlap between `land use' and `land cover', the tasks should be considered `loose' categorisations.}\vspace{-3mm}

\paragraph{Task 1: Land Cover.}
Land cover refers to the biophysical surface characteristics of the Earth (e.g., the type of vegetation or the presence of artificial structures) \cite{diogo2016land, GOMEZ201655}. Each image is labelled with a single broad land cover class, e.g., \textit{forest}, \textit{grassland}, \textit{water} or \textit{building}.

\paragraph{Task 2: Land Use.}
We distinguish land use as the economic and social functions of areas of interest \cite{diogo2016land, comber2008land}. Each image in this task is labelled with a single land use class, which typically offers more granularity than land cover categories -- \textit{artificial structures} could be labelled as \textit{residential} or \textit{industrial}, or \textit{parking lot}, \textit{church} or \textit{storage tank}; categories without a specific socioeconomic function are simply labelled as a land cover class, e.g., \textit{water}.

\paragraph{Task 3: Hierarchical Land Use.}
This task tests the ability to classify land use across different levels of granularity. Each image is annotated with different labels depending on the level. For example, an image of a bridge could be labelled as \textit{transportation area} $\rightarrow$ \textit{highway land} $\rightarrow$ \textit{bridge}.

\end{multicols}
\begin{table*}[t]
\begin{tabular}[!t]{|p{3.75cm}|lllllp{2.9cm}|}
\hline
\multicolumn{1}{|c|}{\multirow{2}{*}{\bf Dataset}} & \multicolumn{1}{p{1.5cm}}{\bf Images in SATIN} & \multicolumn{1}{p{1.5cm}}{\bf Resolution (metres)} & \multicolumn{1}{p{2.0cm}}{\bf Image Width (pixels)} & \multicolumn{1}{c}{\multirow{2}{*}{\bf Region}}  & \multicolumn{1}{c}{\multirow{2}{*}{\bf Classes}} & \multicolumn{1}{c|}{\multirow{2}{*}{\bf Task}} \\
\hline
SAT-4* \cite{basu2015deepsat}                               & 100000  & 1              & 28               & N. America & 4       & Land Cover            \\
SAT-6* \cite{basu2015deepsat}                               & 81000   & 1              & 28               & N. America & 6       & Land Cover            \\
NaSC-TG2 \cite{Zhou2021NaSCTG2}                             & 20000   & 100            & 128              & Global        & 10       & Land Cover            \\
\hline
WHU-RS19 \cite{Xia2010WHURS19, Dai2011WHURS19}                             & 1013    & 0.5            & 600              & -             & 19       & Land Use              \\
RSSCN7 \cite{7272047}                               & 2800    & 1 - 8$^{\ddag}$              & 400              & Global$^{\ddag}$             & 7       & Land Use              \\
RSC11 \cite{zhao2016feature}                                & 1232    & 0.2            & 512              & N. America & 11       & Land Use              \\
SIRI-WHU* \cite{zhao2015dirichlet, zhao2016fisher, zhu2016bag}                            & 2400    & 2              & 200              & Asia          & 12       & Land Use              \\
EuroSAT \cite{helber2018introducing, helber2019eurosat}                              & 27000   & 10             & 64               & Europe        & 10       & Land Use              \\
NWPU-RESISC45  \cite{cheng2017remote}                       & 31500   & 0.2 - 30       & 256              & Global        & 45       & Land Use              \\
PatternNet \cite{zhou2018patternnet}                           & 30400   & 0.06 - 4.7     & 256              & N. America & 38       & Land Use              \\
RSD46-WHU* \cite{long2017accurate, xiao2017high}                           & 117000  & 0.5 - 2        & 256              & Asia$^{\ddag}$          & 46       & Land Use              \\
GID* \cite{GID2020}               & 30000   & 4$^{\ddag}$              & 56; 112; 224     & Asia          & 15       & Land Use              \\
Optimal-31 \cite{wang2018scene}                           & 1860    & 0.2 - 1$^{\ddag}$              & 256              & N. America$^{\ddag}$ & 31       & Land Use              \\
CLRS \cite{s20041226}                                 & 15000   & 0.26 - 8.85    & 256              & Global        & 25       & Land Use              \\
\hline

MillionAID* \cite{long2021creating}                          & 10000 & 0.2 - 153      & 100; 10494      & Global        & 87       & Hier. Land Use \\
RSI-CB256 \cite{li2020RSI-CB}                                   & 24000   & 0.3 - 3        & 256              & Global        & 42       & Hier. Land Use \\
\hline

UCM Land Use$^{\dag}$ \cite{yang2010bag, 8089668}                   & 2100    & 0.3            & 256              & N. America & 17       & Complex Scenes        \\
AID$^{\dag}$ \cite{xia2017aid, hua2019relation}                                  & 10000   & 0.5 - 8        & 600              & Global        & 17       & Complex Scenes        \\
MLRSNet \cite{qi2020mlrsnet}                              & 109161  & 0.1 - 10       & 256              & Global        & 60       & Complex Scenes        \\
MultiScene* \cite{hua2021multiscene}                           & 14000   & 0.3 - 0.6        & 512              & Global        & 36       & Complex Scenes        \\ 
\hline

AWTP*\cite{kaggle_awtp}                & 71504 & 1.5            & 128              & Global     & 2       & Rare Scenes          \\
Post Hurricane* \cite{cao2018deep} & 10000      & 0.46           & 128              & N. America & 2       & Rare Scenes          \\
SISI \cite{kaggle_sisi}                   & 4000    & 3              & 80               & N. America & 2       & Rare Scenes          \\
Canadian Cropland* \cite{jacques2021towards}                   & 14111   & 10             & 64               & N. America & 10       & Rare Scenes          \\
USTC-SmokeRS \cite{ba2019smokenet}                        & 6225    & 1000           & 256              & Global        & 6       & Rare Scenes          \\
\hline

BC Scenes \cite{penatti2015deep}              & 38015   & 2.5$^{\ddag}$            & 64               & S. America & 2       & False Colour Scenes   \\
BCS Scenes \cite{nogueira2016towards}   & 1311    & 5$^{\ddag}$              & 64               & S. America  & 4       & False Colour Scenes  \\
\hline
\end{tabular}
\vspace{-2mm}
\caption{\textbf{SATIN datasets overview.} \textit{Images in SATIN} represents the number of images used in SATIN, not the total number in each dataset. $^{*}$ indicates that a subset of the dataset is used in SATIN. $^{\dag}$ indicates that we use a different set of labels to the original release. $^{\ddag}$ indicates information that was obtained/clarified via correspondence with the authors. More dataset details can be found in App. \ref{section:datasets}. 
}
\label{Table:datasets_overview}

\end{table*}
\begin{multicols}{2}

\paragraph{Task 4: Complex Scenes.}
Given the large fields of view that remote sensing imagery can capture, many scenes contain multiple land use categories. Images used in this task contain one or more land use labels.

\vspace{-0.2cm}
\paragraph{Task 5: Rare Scenes.}
For this task, imagery is classified into rare and more specialised categories that fall outside land cover and land use, e.g., \textit{hurricane damage} or \textit{wind turbines}, or distinguishing between different crop varieties or aerosol classes. Each image is annotated with a single category label.

\vspace{-0.2cm}
\paragraph{Task 6: False Colour Scenes.}
While satellite imagery is arguably most interpretable to humans when in true-colour RGB format, additional insights can be gained by looking at other bands. For this task, we consider two datasets containing false colour images that map the near-infrared, red, green bands to red, green, blue channels. The imagery in these datasets is labelled with a single land use land cover class.
\vspace{2mm}

\subsection{Evaluation}\label{satin:metrics}
We use the accuracy metric to evaluate performance at three levels: \textit{per dataset}, \textit{per task} and \textit{overall}. 
\subsubsection{Dataset Metric}
\textbf{Single Label}.
For the majority of datasets, a single ground-truth label is given per image; the dataset accuracy score is calculated as the fraction of predictions that are correct.
\newline
\textbf{Multiple Labels}.
For images containing multiple ground-truth labels, we compare the $k$ truth labels for a given image with the top $k$ predicted labels; the image classification accuracy is calculated as the Jaccard Index of the lists (the number of labels occurring in both lists forms the numerator, while the number of labels across their union forms the denominator). The dataset accuracy is then calculated as the arithmetic mean over the image accuracies.
\newline
\textbf{Hierarchical Labels}.
Where a hierarchy of labels is given per image, image classification accuracy is determined as the arithmetic mean in prediction accuracies at each level; the dataset accuracy is then calculated as the arithmetic mean over the image accuracies.

\subsubsection{Macro-average Metrics}
We take the geometric mean over the accuracy scores of the datasets for each task; this \textbf{task metric} is used to track progress across tasks and is not included in the overall SATIN metric. Due to an imbalance in the number of datasets per task, we use the geometric mean across datasets (\textit{not tasks}) as the overall \textbf{SATIN metric}.

\renewcommand{\arraystretch}{1}


\section{Experiments}
In this section, we first describe baseline methods evaluated in our work (Sec.~\ref{exp:baselines}) and their inference procedure (Sec.~\ref{exp:inference}). Then, we detail prompt templates used for the aforementioned tasks (Sec.~\ref{exp:prompt})
and report benchmark results (Sec.~\ref{exp:results}).

\subsection{Baselines \& Pretraining}
\label{exp:baselines}
To provide a broad overview of the capability of VL models to interpret RS imagery, we evaluate a range of open-source baselines on our SATIN metadataset. The performance of VL models varies primarily according to three major axes: 
(i) \textbf{Pretraining methodology}---We evaluate the following 8 pretraining methodologies on SATIN: CLIP \cite{radford2021learning}, OpenCLIP \cite{cherti2022reproducible}, ALBEF \cite{li2021align}, BLIP \cite{li2022blip}, BLIP2 \cite{li2023blip}, DeCLIP \cite{li2021supervision}, SLIP \cite{mu2022slip} and CyCLIP \cite{goel2022cyclip}. Although CLIP and OpenCLIP implement the same pretraining strategy (with only minor differences in configuration, e.g., batch-size and learning rate) -- and therefore have no significant differences in performance given identical pretraining data -- we delineate the different implementations for clarity. 
(ii) \textbf{Architecture}---We include a variety of sizes of ResNet- \cite{he2016deep}, ViT- \cite{dosovitskiy2020image} and ConvNeXt- \cite{liu2022convnet} based image encoder backbones. (iii) \textbf{Pretraining data}---Both the quality and volume of pretraining data impacts model performance. We utilise publicly released pretrained checkpoints (details for which can be found in App. \ref{section:methodology}) that have been created with a wide distribution of different pre-training volumes, ranging from 500K to 5.9B examples -- see Table \ref{table:pretraining} for more details. 

\noindent\begin{minipage}[b!]{\columnwidth}
\vspace{2mm}
\raggedright
\begin{tabular}{ll}
\textbf{Training Size} & \textbf{Training Dataset(s)} \\ 
\hline
500K, 1M, 2M, 2.6M & CC3M \cite{sharma2018conceptual} \\
3M & CC3M \cite{sharma2018conceptual} + CC12M \cite{changpinyo2021conceptual} \\
14M & See \cite{li2021align} for details \\
15M & YFCC \cite{thomee2016yfcc100m} \\
88M & DeCLIP Full Data \cite{li2021supervision}  \\
129M & See \cite{li2022blip} for details \\
400M & CLIP-WIT \cite{radford2021learning}  \\
400M & LAION-400M \cite{schuhmann2021laion} \\
2B & LAION-2B \cite{schuhmann2022laion}  \\
5.9B & LAION-5B \cite{schuhmann2022laion} \\
\hline
\end{tabular}
\vspace{-0.15cm}
\captionof{table}{\textbf{Details of pretraining datasets.} \textit{Training Size} denotes the number of image-text examples in the respective datasets.} 
\label{table:pretraining}
\end{minipage}

\begin{table*}[!th]
    \small
    \centering
    \begin{tabular}{l l l | c c c c c c c}
    \hline
    \multicolumn{3}{c |}{\bf \textit{Model}}  & \multicolumn{7}{c}{\bf \textit{Zero-Shot Classification Accuracy}}
    \\
    \multicolumn{1}{l}{\textbf{Method}} & \multicolumn{1}{l}{\bf Backbone} & \multicolumn{1}{l |}{\bf Pretraining $\uparrow$} & \multicolumn{1}{c}{\bf Task 1} & \multicolumn{1}{c}{\bf Task 2}  & \multicolumn{1}{c}{\bf Task 3} & \multicolumn{1}{c}{\bf Task 4}  & \multicolumn{1}{c}{\bf Task 5} & \multicolumn{1}{c}{\bf Task 6} & \multicolumn{1}{c}{\bf SATIN}
    \\  
    \hline

     CyCLIP \cite{goel2022cyclip} & RN50 & 3M & 0.20 & 0.22 & 0.27 & 0.37 & 0.28 & 0.29 & 0.25 \\
     ALBEF \cite{li2021align} & ViT-B/16 & 14M & 0.34 & 0.37 & 0.33 & 0.38 & 0.39 & 0.51 & 0.38 \\
     SLIP \cite{mu2022slip} & ViT-B/32 & 15M & 0.30 & 0.22 & 0.19 & 0.29 & 0.36 & 0.22 & 0.26 \\
     DeCLIP \cite{li2021supervision} & ViT-B/32 & 88M & 0.43 & 0.42 & 0.37 & 0.46 & 0.32 & \textbf{0.58} & 0.41 \\
     BLIP \cite{li2022blip} & ViT-B/16 & 129M & 0.37 & 0.50 & 0.45 & 0.50 & 0.37 & 0.46 & 0.45 \\
     BLIP2 \cite{li2023blip} & ViT-G/14 & 129M & 0.47 & 0.55 & 0.50 & \textbf{0.51} &\textbf{ 0.50} & 0.21 & 0.48 \\
     CLIP \cite{radford2021learning} & ViT-L/14@336px & 400M & \textbf{0.53} & 0.59 & \textbf{0.51} & 0.48 & 0.44 & 0.35 & 0.51 \\
     OpenCLIP \cite{cherti2022reproducible} & ViT-G/14 & 2B & 0.52 & \textbf{0.63} & 0.50 & 0.50 & 0.44 & 0.31 & \textbf{0.52} \\
            
    \hline
    \end{tabular}
    \vspace{-0.25cm}
    \caption{\textbf{Baseline zero-shot performance on the SATIN benchmark.} For each method, we evaluate the available model with the highest capacity backbone and largest volume of pretraining data. We find that the best model achieves an overall score of just 52\% on SATIN.
    }
    \label{Table:best_8_results}
\end{table*}

\subsection{Inference}\label{exp:inference}
\noindent\textbf{Steps.} Although the baseline methods differ in their pretraining strategy, the steps taken by each model during inference can be generalised as the following: (1) Preprocess input images and text; (2) Extract image and text features; (3) Normalise features; (4) Compute similarity; (5) Determine top predicted class(es). 
See App. \ref{appendix:pseudocode} for pseudocode. 

\noindent\textbf{Compute.}
The inference time required to evaluate SATIN varies considerably depending on the model but remains practical for researchers with access to a single GPU. For instance, the BLIP model takes approximately 30 minutes to evaluate across the full SATIN benchmark with one A100 GPU.  
A more detailed analysis can be found in App. \ref{appendix:compute}, including inference times per dataset, per model; as well as GPU memory.

\subsection{Prompt Engineering}\label{exp:prompt}
Rather than tuning to particular datasets, we aim to provide a general overview of capability and therefore use generic prompt templates. For the majority of tasks, we use: ``\textit{a satellite photo of CLS}." and ``\textit{an aerial photo of CLS}.", where CLS represents the target classes. To ensure grammatical sense, we use ``\textit{a satellite photo containing CLS}." and ``\textit{an aerial photo containing CLS}." for the Complex Scenes task where there are one or more labels per image. To contextualise the different imagery format used in the False-Colour Scenes task, we add \textit{false-colour} to the prompt. We take the text embedding for each target class as the average over the two prompt templates.

\subsection{Benchmarking Results}
\label{exp:results}
\noindent\textbf{Zero-shot classification results for different methods.}
Table \ref{Table:best_8_results} shows the zero-shot classification results for each of the 8 methods, using checkpoints corresponding to the highest capacity backbones and largest volumes of pretraining images. Considering the overall SATIN metric, a general trend can be observed that performance increases with higher capacity backbones and exposure to a greater number of pretraining image-text pairs. This is in line with existing literature that has compared different backbones and levels of pretraining, e.g., \cite{radford2021learning, cherti2022reproducible}.
The performance across the different tasks largely follows the same trend; a notable example, however, is Task 6, which sees the larger models struggling relative to those with lower capacity backbones and smaller-scale pretraining, such as the ALBEF and DeCLIP implementations. The fact that Task 6 is comprised of false-colour imagery perhaps favours models that exhibit less overfitting to natural images. Given the variation across pretraining data and backbones, this does not constitute a direct comparison of the different methods. 
\textbf{Key insight---}\textit{In terms of overall performance, even the largest capacity models barely surpass 50\% classification accuracy, conveying that SATIN represents a challenging benchmark, suitably placed with sufficient capability in the models to significantly beat the chance score while leaving plenty of room for improvement. }

\begin{figure*}[!th]
  \centering
  \includegraphics[width=0.90\textwidth]{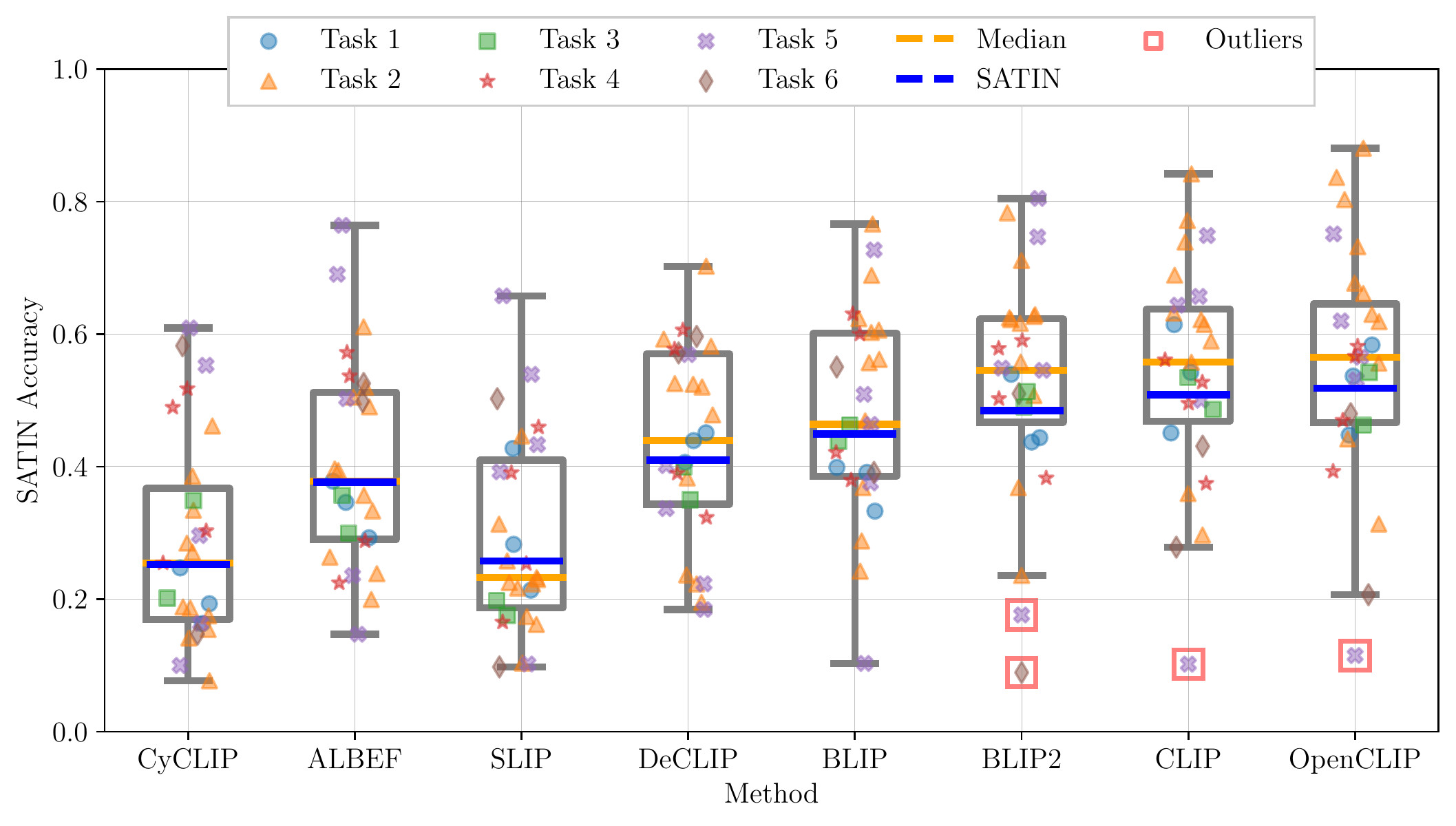}
  \vspace{-0.25cm}
  \caption{\textbf{SATIN task performance distribution for the models in Table \ref{Table:best_8_results}.} A small amount of random horizontal jitter has been arbitrarily added to the scatter points to improve readability. The lowest-scoring purple cross represents the challenging Canadian Cropland dataset.} 
  \vspace{-0.3cm}
  \label{Figure:boxplots}
\end{figure*}


\noindent\textbf{Consistency across datasets.} Figure \ref{Figure:boxplots} provides an overview of the consistency of each method across the different datasets in the SATIN benchmark. \textbf{Key insight---}\textit{We observe a wide distribution of performance across the datasets, spanning at least 40\% for each method.} This spread gets larger with increasing performance as the higher capacity models attain higher scores on the easier datasets but show little improvement on a few challenging ones, such as those highlighted as outliers. An especially challenging dataset is the Canadian Cropland dataset, which contains images of 10 different crop species that are difficult for humans to distinguish (though there is sufficient information in the imagery for supervised methods to attain far beyond the chance score of 0.1 \cite{jacques2021towards}). On the other hand, some datasets such as WHU-RS19 are significantly easier, with scores achieved that approach the 90\% mark.

\begin{figure*}[!th]
  \centering
  \includegraphics[width=\textwidth]{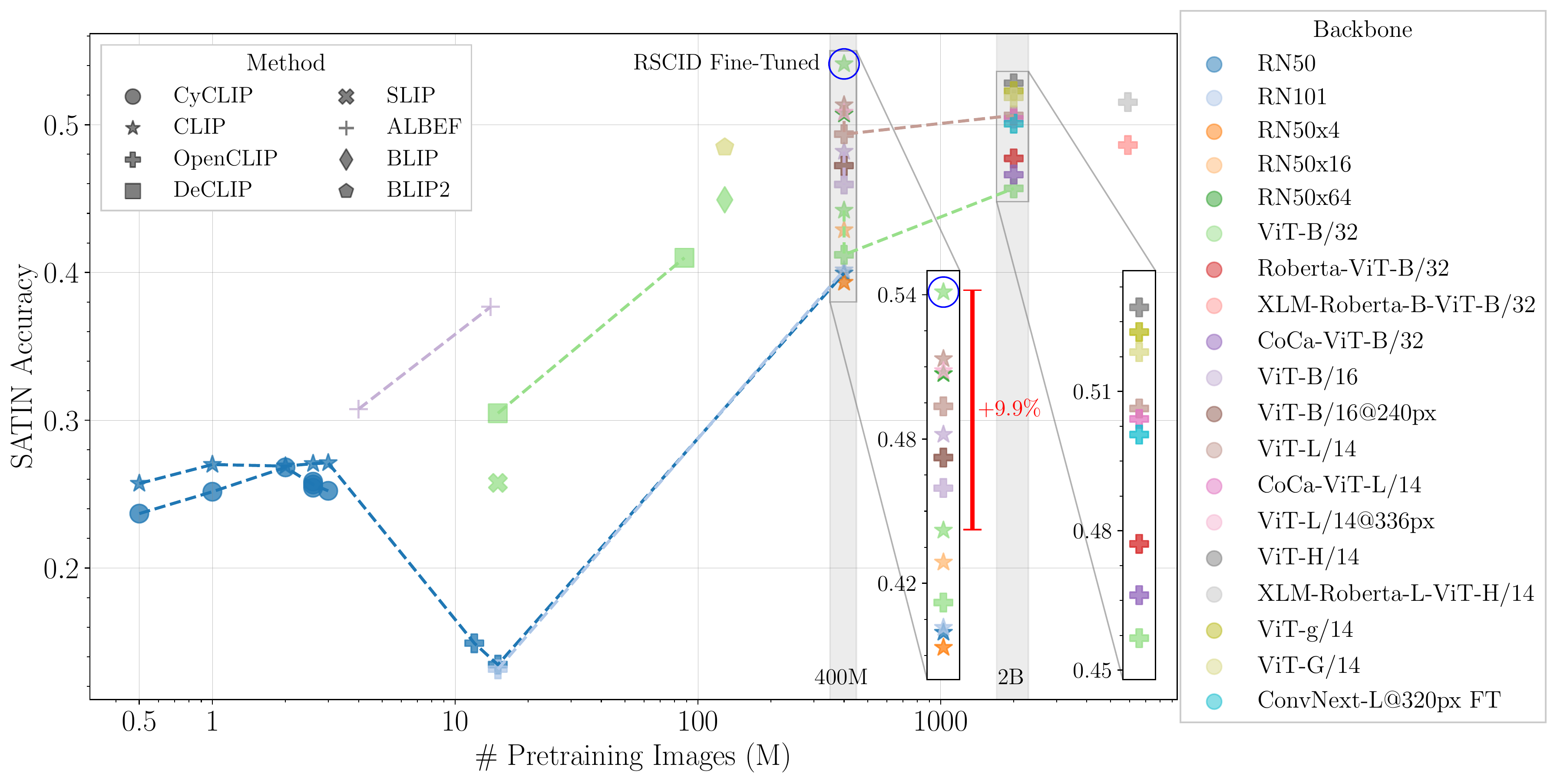}
  \vspace{-0.75cm}
  \caption{\textbf{SATIN zero-shot performance for a broad range of VL models.} We delineate the performance on SATIN of different models by denoting \textit{methods} with different symbols, \textit{backbones} with different colours, and \textit{number of pretraining images} across the x-axis. Models with the same method and backbone are linked with dashed lines and for clarity, insets are added of the crowded regions. In general, we observe increasing SATIN accuracy with the volume of pretraining data. Of the models we benchmark, we find that a ViT-B/32 CLIP model fine-tuned with a small volume of RS data achieves the highest score.}
  \label{fig:scatter}
\end{figure*}

\noindent\textbf{SATIN performance wrt. pretraining.}
In addition to the models evaluated in Table \ref{Table:best_8_results}, we evaluate a wider profile of method-backbone combinations at different pretrainined checkpoints on SATIN, covering a total of 20 different image encoder architectures and 15 different pretraining datasets; we also include a model fine-tuned on RS data. A detailed breakdown of the per dataset scores for each model can be found in App. \ref{section:results}. In Figure \ref{fig:scatter}, we distill this performance information into a comparison of different models through the lens of pretraining image volume.
\textbf{Key insight---}\textit{Consideration of the overall distribution of points shows a general trend of increasing accuracy against the SATIN benchmark for models pretrained with a higher volume of images.}
The dashed lines link models with matching methods \textit{and} backbones, allowing for a more direct comparison of the effect of pretraining data. In the majority of cases, performance is correlated with the number of pretraining examples -- this is especially apparent for the ALBEF (light purple crosses) and DeCLIP (light green squares) models. This can also be seen for the OpenCLIP models in the 400M and 2B regions. Despite having equivalent pretraining strategies, the CLIP 400M models noticeably outperform the corresponding OpenCLIP 400M models -- as shown by the light pink, green and brown vertical dashed lines in the 400M region (representing the ViT-L/14, ViT-B/16 and ViT-B/32 backbones, respectively) -- reflecting the difference in the quality of their distinct but equally sized pretraining datasets.
Pretraining quality over quantity is also observable for the RN50 CLIP models in the lower data regime, where the 3M model outperforms the 12M model, which in turn outperforms the 15M model (see blue dashed line). Within the 400M and 2B regions, the higher capacity ViT models attain stronger performances, as expected.


\noindent\textbf{Fine-tuning results.} Given the disparity between the mostly natural images used during pretraining and in-domain RS images, it is expected that fine-tuning a model with even a relatively small volume of RS images would yield significant performance increases. We evaluate a CLIP ViT-B/32 (400M) model fine-tuned on $\sim$13K multi-captioned RS images\footnote{\url{https://huggingface.co/flax-community/clip-rsicd-v2}} from the RSICD \cite{lu2017exploring}, Sydney-captions \cite{qu2016cits} and UCM-captions \cite{qu2016cits} datasets. Despite the fact that these models have the lowest capacity ViT image encoders, the small-scale fine-tuning results in them attaining the highest scores of all the models evaluated, outperforming those with larger image encoders and volumes of pretraining data. \textbf{Key insight---}\textit{By comparing with the equivalent model without fine-tuning we find that the absolute benefits of this fine-tuning is a performance improvement of +9.9\%.} The data used during fine-tuning does include imagery (albeit with different labels) from one of the SATIN datasets, however, this represents just 1 of 27 datasets in the benchmark.

\section{Broader Impacts}
 
There are many positive applications to research involving RS data, many of which are in pursuit of achieving the UN's SGDs\footnote{\url{https://sdgs.un.org/goals}}. Due to the nature of RS data, there is also potential for its usage in areas that are less than unanimously accepted, such as surveillance.

To minimise bias and promote robustness, we aimed to maximise geographic diversity and representation within SATIN by casting a wide net and considering a broad distribution of potential constituent datasets. Despite this intention, we are still limited by the set of possible datasets that are publicly available. Although we include many globally distributed datasets, the majority of the seminal and older works tend to be limited to specific countries or continents, typically North America, Europe, and China. Furthermore, the datasets with global coverage tend to be annotated using crowdsourced platforms such as OpenStreetMap\footnote{\url{https://www.openstreetmap.org/}} and therefore disproportionately represent locations with more users and annotations.
\section{Limitations}

As with many datasets involving complex scenes with large fields of view, issues arise when assigning a single classification label. While we mitigate this by including multiple labels where possible, we do not take any measures to remove background. Therefore, it might be impossible to attain 100\% accuracy on some of the datasets.

In SATIN, we only use optical imagery (RGB + NIR), though some RS platforms capture imagery in other parts of the electromagnetic spectrum, such as the radio spectrum. While additional information can be garnered from these extra bands, we limit the scope of our evaluation to include optical imagery as it is closer to the distribution of images used during pre-training.
\section{Conclusion}

We introduce SATIN, a remote sensing image classification metadataset curated from 27 existing datasets covering a wide range of resolutions, image sizes and geographic regions, grouped into 6 different tasks. 
We benchmark the zero-shot classification capabilities of a broad profile of over 40 vision-language model baselines -- comprising different backbones, methods and pretraining data -- against SATIN. 
We observe that SATIN proves a challenge to even the highest capacity models trained on billions of natural images, which attain an accuracy score of just over 50\%. We find a wide variation in performance on the constituent datasets within SATIN, ranging from slightly above chance score to approaching 90\%. We evaluate a small number of fine-tuned models against SATIN and find significant performance increases after exposure to a relatively low number of in-domain images, though plenty of room for improvement remains. We release SATIN and provide a public leaderboard to track progress in the development of vision-language models for remote sensing image interpretation, a domain with a multitude of socially beneficial applications. 

\section*{Acknowledgements}
We thank the authors of the original datasets included in SATIN for their help during the curation of our benchmark. We also thank Gyungin Shin and Vishaal Udandarao for their thoughtful comments when reviewing this manuscript.

This work was supported by the UKRI Centre for Doctoral Training in Application of Artificial Intelligence to the study of Environmental Risks (reference EP/S022961/1), an Isaac Newton Trust grant, an EPSRC HPC grant, the Hong Kong Research Grant Council - Early Career Scheme (Grant No. 27208022), and HKU Seed Fund for Basic Research. Samuel would like to acknowledge the support of Z. Novak and N. Novak in enabling his contribution.
{\small
\bibliographystyle{ieee_fullname}
\bibliography{egbib}

\begin{thebibliography}{10}\itemsep=-1pt

\bibitem{TFDS}
{TensorFlow Datasets}, a collection of ready-to-use datasets.
\newblock \url{https://www.tensorflow.org/datasets}.

\bibitem{ba2019smokenet}
Rui Ba, Chen Chen, Jing Yuan, Weiguo Song, and Siuming Lo.
\newblock Smokenet: Satellite smoke scene detection using convolutional neural
  network with spatial and channel-wise attention.
\newblock {\em Remote Sensing}, 11(14):1702, 2019.

\bibitem{bastani2022satlas}
Favyen Bastani, Piper Wolters, Ritwik Gupta, Joe Ferdinando, and Aniruddha
  Kembhavi.
\newblock Satlas: A large-scale, multi-task dataset for remote sensing image
  understanding.
\newblock {\em arXiv preprint arXiv:2211.15660}, 2022.

\bibitem{basu2015deepsat}
Saikat Basu, Sangram Ganguly, Supratik Mukhopadhyay, Robert DiBiano, Manohar
  Karki, and Ramakrishna Nemani.
\newblock Deepsat: a learning framework for satellite imagery.
\newblock In {\em Proceedings of the 23rd SIGSPATIAL international conference
  on advances in geographic information systems}, pages 1--10, 2015.

\bibitem{boori2015monitoring}
Mukesh~Singh Boori, Maik Netzband, Komal Choudhary, and V{\'\i}t
  Vo{\v{z}}en{\'\i}lek.
\newblock Monitoring and modeling of urban sprawl through remote sensing and
  gis in kuala lumpur, malaysia.
\newblock {\em Ecological Processes}, 4(1):1--10, 2015.

\bibitem{bornschein2022nevis}
Jorg Bornschein, Alexandre Galashov, Ross Hemsley, Amal Rannen-Triki, Yutian
  Chen, Arslan Chaudhry, Xu~Owen He, Arthur Douillard, Massimo Caccia, Qixuang
  Feng, et~al.
\newblock Nevis'22: A stream of 100 tasks sampled from 30 years of computer
  vision research.
\newblock {\em arXiv preprint arXiv:2211.11747}, 2022.

\bibitem{cao2018deep}
Quoc~Dung Cao and Youngjun Choe.
\newblock Deep learning based damage detection on post-hurricane satellite
  imagery.
\newblock {\em arXiv preprint arXiv:1807.01688}, 2018.

\bibitem{changpinyo2021conceptual}
Soravit Changpinyo, Piyush Sharma, Nan Ding, and Radu Soricut.
\newblock Conceptual 12m: Pushing web-scale image-text pre-training to
  recognize long-tail visual concepts.
\newblock In {\em Proceedings of the IEEE/CVF Conference on Computer Vision and
  Pattern Recognition}, pages 3558--3568, 2021.

\bibitem{8089668}
Bindita Chaudhuri, Begüm Demir, Subhasis Chaudhuri, and Lorenzo Bruzzone.
\newblock Multilabel remote sensing image retrieval using a semisupervised
  graph-theoretic method.
\newblock {\em IEEE Transactions on Geoscience and Remote Sensing},
  56(2):1144--1158, 2018.

\bibitem{cheng2017remote}
Gong Cheng, Junwei Han, and Xiaoqiang Lu.
\newblock Remote sensing image scene classification: Benchmark and state of the
  art.
\newblock {\em Proceedings of the IEEE}, 105(10):1865--1883, 2017.

\bibitem{cherti2022reproducible}
Mehdi Cherti, Romain Beaumont, Ross Wightman, Mitchell Wortsman, Gabriel
  Ilharco, Cade Gordon, Christoph Schuhmann, Ludwig Schmidt, and Jenia Jitsev.
\newblock Reproducible scaling laws for contrastive language-image learning.
\newblock {\em arXiv preprint arXiv:2212.07143}, 2022.

\bibitem{christie2018functional}
Gordon Christie, Neil Fendley, James Wilson, and Ryan Mukherjee.
\newblock Functional map of the world.
\newblock In {\em Proceedings of the IEEE Conference on Computer Vision and
  Pattern Recognition}, pages 6172--6180, 2018.

\bibitem{cihlar2000land}
J Cihlar.
\newblock Land cover mapping of large areas from satellites: status and
  research priorities.
\newblock {\em International journal of remote sensing}, 21(6-7):1093--1114,
  2000.

\bibitem{comber2008land}
Alexis~J Comber.
\newblock Land use or land cover?
\newblock {\em Journal of Land Use Science}, 3(4):199--201, 2008.

\bibitem{congalton2014global}
Russell~G Congalton, Jianyu Gu, Kamini Yadav, Prasad Thenkabail, and Mutlu
  Ozdogan.
\newblock Global land cover mapping: A review and uncertainty analysis.
\newblock {\em Remote Sensing}, 6(12):12070--12093, 2014.

\bibitem{cornebise2022open}
Julien Cornebise, Ivan Or{\v{s}}oli{\'c}, and Freddie Kalaitzis.
\newblock Open high-resolution satellite imagery: The worldstrat dataset--with
  application to super-resolution.
\newblock {\em arXiv preprint arXiv:2207.06418}, 2022.

\bibitem{Dai2011WHURS19}
Dengxin Dai and Wen Yang.
\newblock Satellite image classification via two-layer sparse coding with
  biased image representation.
\newblock {\em IEEE Transactions on Geoscience and Remote Sensing},
  8(1):173--176, 2011.

\bibitem{DIMITROVSKI202318}
Ivica Dimitrovski, Ivan Kitanovski, Dragi Kocev, and Nikola Simidjievski.
\newblock Current trends in deep learning for earth observation: An open-source
  benchmark arena for image classification.
\newblock {\em ISPRS Journal of Photogrammetry and Remote Sensing}, 197:18--35,
  2023.

\bibitem{diogo2016land}
Vasco Diogo and Eric Koomen.
\newblock Land cover and land use indicators: review of available data.
\newblock {\em OECD green growth Paper}, 2016.

\bibitem{dosovitskiy2020image}
Alexey Dosovitskiy, Lucas Beyer, Alexander Kolesnikov, Dirk Weissenborn,
  Xiaohua Zhai, Thomas Unterthiner, Mostafa Dehghani, Matthias Minderer, Georg
  Heigold, Sylvain Gelly, et~al.
\newblock An image is worth 16x16 words: Transformers for image recognition at
  scale.
\newblock {\em arXiv preprint arXiv:2010.11929}, 2020.

\bibitem{dubovik}
Oleg Dubovik, Gregory~L. Schuster, Feng Xu, Yongxiang Hu, Hartmut Bösch,
  Jochen Landgraf, and Zhengqiang Li.
\newblock Grand challenges in satellite remote sensing.
\newblock {\em Frontiers in Remote Sensing}, 2, 2021.

\bibitem{fritz2010comparison}
Steffen Fritz, Linda See, and Felix Rembold.
\newblock Comparison of global and regional land cover maps with statistical
  information for the agricultural domain in africa.
\newblock {\em International Journal of Remote Sensing}, 31(9):2237--2256,
  2010.

\bibitem{eval2021harness}
Leo Gao, Jonathan Tow, Stella Biderman, Sid Black, Anthony DiPofi, Charles
  Foster, Laurence Golding, Jeffrey Hsu, Kyle McDonell, Niklas Muennighoff,
  Jason Phang, Laria Reynolds, Eric Tang, Anish Thite, Ben Wang, Kevin Wang,
  and Andy Zou.
\newblock A framework for few-shot language model evaluation, Sept. 2021.

\bibitem{goel2022cyclip}
Shashank Goel, Hritik Bansal, Sumit Bhatia, Ryan~A Rossi, Vishwa Vinay, and
  Aditya Grover.
\newblock Cyclip: Cyclic contrastive language-image pretraining.
\newblock {\em arXiv preprint arXiv:2205.14459}, 2022.

\bibitem{grinand2013estimating}
Clovis Grinand, Fety Rakotomalala, Val{\'e}ry Gond, Romuald Vaudry, Martial
  Bernoux, and Ghislain Vieilledent.
\newblock Estimating deforestation in tropical humid and dry forests in
  madagascar from 2000 to 2010 using multi-date landsat satellite images and
  the random forests classifier.
\newblock {\em Remote Sensing of Environment}, 139:68--80, 2013.

\bibitem{GOMEZ201655}
Cristina Gómez, Joanne~C. White, and Michael~A. Wulder.
\newblock Optical remotely sensed time series data for land cover
  classification: A review.
\newblock {\em ISPRS Journal of Photogrammetry and Remote Sensing}, 116:55--72,
  2016.

\bibitem{kaggle_sisi}
Robert Hammell.
\newblock Ships in satellite imagery.
\newblock
  \url{https://www.kaggle.com/datasets/rhammell/ships-in-satellite-imagery},
  2018.

\bibitem{he2016deep}
Kaiming He, Xiangyu Zhang, Shaoqing Ren, and Jian Sun.
\newblock Deep residual learning for image recognition.
\newblock In {\em Proceedings of the IEEE conference on computer vision and
  pattern recognition}, pages 770--778, 2016.

\bibitem{helber2018introducing}
Patrick Helber, Benjamin Bischke, Andreas Dengel, and Damian Borth.
\newblock Introducing eurosat: A novel dataset and deep learning benchmark for
  land use and land cover classification.
\newblock In {\em IGARSS 2018-2018 IEEE International Geoscience and Remote
  Sensing Symposium}, pages 204--207. IEEE, 2018.

\bibitem{helber2019eurosat}
Patrick Helber, Benjamin Bischke, Andreas Dengel, and Damian Borth.
\newblock Eurosat: A novel dataset and deep learning benchmark for land use and
  land cover classification.
\newblock {\em IEEE Journal of Selected Topics in Applied Earth Observations
  and Remote Sensing}, 2019.

\bibitem{hua2021multiscene}
Y. Hua, L. Mou, P. Jin, and X.~X. Zhu.
\newblock Multiscene: A large-scale dataset and benchmark for multi-scene
  recognition in single aerial images.
\newblock {\em IEEE Transactions on Geoscience and Remote Sensing}, in press.

\bibitem{hua2019relation}
Yuansheng Hua, Lichao Mou, and Xiao~Xiang Zhu.
\newblock Relation network for multi-label aerial image classification.
\newblock {\em IEEE Transactions on Geoscience and Remote Sensing},
  DOI:10.1109/TGRS.2019.2963364.

\bibitem{jacques2021towards}
Amanda A~Boatswain Jacques, Abdoulaye~Banir{\'e} Diallo, and Etienne Lord.
\newblock Towards the creation of a canadian land-use dataset for agricultural
  land classification.
\newblock In {\em 42nd Canadian Symposium on Remote Sensing: Understanding Our
  World: Remote Sensing for a Sustainable Future}, 2021.

\bibitem{jia2021scaling}
Chao Jia, Yinfei Yang, Ye Xia, Yi-Ting Chen, Zarana Parekh, Hieu Pham, Quoc Le,
  Yun-Hsuan Sung, Zhen Li, and Tom Duerig.
\newblock Scaling up visual and vision-language representation learning with
  noisy text supervision.
\newblock In {\em International Conference on Machine Learning}, pages
  4904--4916. PMLR, 2021.

\bibitem{KERR2003299}
Jeremy~T. Kerr and Marsha Ostrovsky.
\newblock From space to species: ecological applications for remote sensing.
\newblock {\em Trends in Ecology \& Evolution}, 18(6):299--305, 2003.

\bibitem{kim2020new}
Jhoon Kim, Ukkyo Jeong, Myoung-Hwan Ahn, Jae~H Kim, Rokjin~J Park, Hanlim Lee,
  Chul~Han Song, Yong-Sang Choi, Kwon-Ho Lee, Jung-Moon Yoo, et~al.
\newblock New era of air quality monitoring from space: Geostationary
  environment monitoring spectrometer (gems).
\newblock {\em Bulletin of the American Meteorological Society},
  101(1):E1--E22, 2020.

\bibitem{lhoest-etal-2021-datasets}
Quentin Lhoest, Albert Villanova~del Moral, Yacine Jernite, Abhishek Thakur,
  Patrick von Platen, Suraj Patil, Julien Chaumond, Mariama Drame, Julien Plu,
  Lewis Tunstall, Joe Davison, Mario {\v{S}}a{\v{s}}ko, Gunjan Chhablani,
  Bhavitvya Malik, Simon Brandeis, Teven Le~Scao, Victor Sanh, Canwen Xu,
  Nicolas Patry, Angelina McMillan-Major, Philipp Schmid, Sylvain Gugger,
  Cl{\'e}ment Delangue, Th{\'e}o Matussi{\`e}re, Lysandre Debut, Stas Bekman,
  Pierric Cistac, Thibault Goehringer, Victor Mustar, Fran{\c{c}}ois Lagunas,
  Alexander Rush, and Thomas Wolf.
\newblock Datasets: A community library for natural language processing.
\newblock In {\em Proceedings of the 2021 Conference on Empirical Methods in
  Natural Language Processing: System Demonstrations}, pages 175--184, Online
  and Punta Cana, Dominican Republic, Nov. 2021. Association for Computational
  Linguistics.

\bibitem{li2022elevater}
Chunyuan Li, Haotian Liu, Liunian~Harold Li, Pengchuan Zhang, Jyoti Aneja,
  Jianwei Yang, Ping Jin, Yong~Jae Lee, Houdong Hu, Zicheng Liu, et~al.
\newblock Elevater: A benchmark and toolkit for evaluating language-augmented
  visual models.
\newblock {\em arXiv preprint arXiv:2204.08790}, 2022.

\bibitem{li2020RSI-CB}
Haifeng Li, Xin Dou, Chao Tao, Zhixiang Wu, Jie Chen, Jian Peng, Min Deng, and
  Ling Zhao.
\newblock Rsi-cb: A large-scale remote sensing image classification benchmark
  using crowdsourced data.
\newblock {\em Sensors}, 20(6):1594, 2020.

\bibitem{s20041226}
Haifeng Li, Hao Jiang, Xin Gu, Jian Peng, Wenbo Li, Liang Hong, and Chao Tao.
\newblock Clrs: Continual learning benchmark for remote sensing image scene
  classification.
\newblock {\em Sensors}, 20(4), 2020.

\bibitem{li2023blip}
Junnan Li, Dongxu Li, Silvio Savarese, and Steven Hoi.
\newblock Blip-2: Bootstrapping language-image pre-training with frozen image
  encoders and large language models.
\newblock {\em arXiv preprint arXiv:2301.12597}, 2023.

\bibitem{li2022blip}
Junnan Li, Dongxu Li, Caiming Xiong, and Steven Hoi.
\newblock Blip: Bootstrapping language-image pre-training for unified
  vision-language understanding and generation.
\newblock In {\em International Conference on Machine Learning}, pages
  12888--12900. PMLR, 2022.

\bibitem{li2021align}
Junnan Li, Ramprasaath Selvaraju, Akhilesh Gotmare, Shafiq Joty, Caiming Xiong,
  and Steven Chu~Hong Hoi.
\newblock Align before fuse: Vision and language representation learning with
  momentum distillation.
\newblock {\em Advances in neural information processing systems},
  34:9694--9705, 2021.

\bibitem{li2021supervision}
Yangguang Li, Feng Liang, Lichen Zhao, Yufeng Cui, Wanli Ouyang, Jing Shao,
  Fengwei Yu, and Junjie Yan.
\newblock Supervision exists everywhere: A data efficient contrastive
  language-image pre-training paradigm.
\newblock {\em arXiv preprint arXiv:2110.05208}, 2021.

\bibitem{liang2022holistic}
Percy Liang, Rishi Bommasani, Tony Lee, Dimitris Tsipras, Dilara Soylu,
  Michihiro Yasunaga, Yian Zhang, Deepak Narayanan, Yuhuai Wu, Ananya Kumar,
  et~al.
\newblock Holistic evaluation of language models.
\newblock {\em arXiv preprint arXiv:2211.09110}, 2022.

\bibitem{liu2022convnet}
Zhuang Liu, Hanzi Mao, Chao-Yuan Wu, Christoph Feichtenhofer, Trevor Darrell,
  and Saining Xie.
\newblock A convnet for the 2020s.
\newblock In {\em Proceedings of the IEEE/CVF Conference on Computer Vision and
  Pattern Recognition}, pages 11976--11986, 2022.

\bibitem{long2017accurate}
Yang Long, Yiping Gong, Zhifeng Xiao, and Qing Liu.
\newblock Accurate object localization in remote sensing images based on
  convolutional neural networks.
\newblock {\em IEEE Transactions on Geoscience and Remote Sensing},
  55(5):2486--2498, 2017.

\bibitem{long2021creating}
Yang Long, Gui-Song Xia, Shengyang Li, Wen Yang, Michael~Ying Yang, Xiao~Xiang
  Zhu, Liangpei Zhang, and Deren Li.
\newblock On creating benchmark dataset for aerial image interpretation:
  Reviews, guidances, and million-aid.
\newblock {\em IEEE Journal of selected topics in applied earth observations
  and remote sensing}, 14:4205--4230, 2021.

\bibitem{lu2017exploring}
Xiaoqiang Lu, Binqiang Wang, Xiangtao Zheng, and Xuelong Li.
\newblock Exploring models and data for remote sensing image caption
  generation.
\newblock {\em IEEE Transactions on Geoscience and Remote Sensing},
  56(4):2183--2195, 2018.

\bibitem{makynen2020satellite}
Marko M{\"a}kynen, Jari Haapala, Giuseppe Aulicino, Beena Balan-Sarojini,
  Magdalena Balmaseda, Alexandru Gegiuc, Fanny Girard-Ardhuin, Stefan
  Hendricks, Georg Heygster, Larysa Istomina, et~al.
\newblock Satellite observations for detecting and forecasting sea-ice
  conditions: A summary of advances made in the spices project by the eu’s
  horizon 2020 programme.
\newblock {\em Remote Sensing}, 12(7):1214, 2020.

\bibitem{mitchell2017current}
Anthea~L Mitchell, Ake Rosenqvist, and Brice Mora.
\newblock Current remote sensing approaches to monitoring forest degradation in
  support of countries measurement, reporting and verification (mrv) systems
  for redd+.
\newblock {\em Carbon balance and management}, 12(1):1--22, 2017.

\bibitem{mu2022slip}
Norman Mu, Alexander Kirillov, David Wagner, and Saining Xie.
\newblock Slip: Self-supervision meets language-image pre-training.
\newblock In {\em Computer Vision--ECCV 2022: 17th European Conference, Tel
  Aviv, Israel, October 23--27, 2022, Proceedings, Part XXVI}, pages 529--544.
  Springer, 2022.

\bibitem{nogueira2016towards}
Keiller Nogueira, Jefersson~A Dos~Santos, Tamires Fornazari, Thiago
  Sanna~Freire Silva, Leonor~Patricia Morellato, and Ricardo da~S Torres.
\newblock Towards vegetation species discrimination by using data-driven
  descriptors.
\newblock In {\em 2016 9th IAPR Workshop on Pattern Recogniton in Remote
  Sensing (PRRS)}, pages 1--6. Ieee, 2016.

\bibitem{penatti2015deep}
Ot{\'a}vio~AB Penatti, Keiller Nogueira, and Jefersson~A Dos~Santos.
\newblock Do deep features generalize from everyday objects to remote sensing
  and aerial scenes domains?
\newblock In {\em Proceedings of the IEEE conference on computer vision and
  pattern recognition workshops}, pages 44--51, 2015.

\bibitem{qi2020mlrsnet}
Xiaoman Qi, Panpan Zhu, Yuebin Wang, Liqiang Zhang, Junhuan Peng, Mengfan Wu,
  Jialong Chen, Xudong Zhao, Ning Zang, and P~Takis Mathiopoulos.
\newblock Mlrsnet: A multi-label high spatial resolution remote sensing dataset
  for semantic scene understanding.
\newblock {\em ISPRS Journal of Photogrammetry and Remote Sensing},
  169:337--350, 2020.

\bibitem{qu2016cits}
Bo Qu, Xuelong Li, Dacheng Tao, and Xiaoqiang Lu.
\newblock Deep semantic understanding of high resolution remote sensing image.
\newblock In {\em 2016 International Conference on Computer, Information and
  Telecommunication Systems (CITS)}, pages 1--5, 2016.

\bibitem{radford2021learning}
Alec Radford, Jong~Wook Kim, Chris Hallacy, Aditya Ramesh, Gabriel Goh,
  Sandhini Agarwal, Girish Sastry, Amanda Askell, Pamela Mishkin, Jack Clark,
  et~al.
\newblock Learning transferable visual models from natural language
  supervision.
\newblock In {\em International Conference on Machine Learning}, pages
  8748--8763. PMLR, 2021.

\bibitem{rogan2004remote}
John Rogan and DongMei Chen.
\newblock Remote sensing technology for mapping and monitoring land-cover and
  land-use change.
\newblock {\em Progress in planning}, 61(4):301--325, 2004.

\bibitem{kaggle_awtp}
Airbus DS~GEO S.A.
\newblock Airbus wind turbine patches.
\newblock
  \url{https://www.kaggle.com/datasets/airbusgeo/airbus-wind-turbines-patches},
  2021.

\bibitem{saah_landcover_examples}
David Saah, Karis Tenneson, Mir Matin, Kabir Uddin, Peter Cutter, Ate
  Poortinga, Quyen~H. Nguyen, Matthew Patterson, Gary Johnson, Kel Markert,
  Africa Flores, Eric Anderson, Amanda Weigel, Walter~L. Ellenberg, Radhika
  Bhargava, Aekkapol Aekakkararungroj, Biplov Bhandari, Nishanta Khanal, Ian~W.
  Housman, Peter Potapov, Alexandra Tyukavina, Paul Maus, David Ganz, Nicholas
  Clinton, and Farrukh Chishtie.
\newblock Land cover mapping in data scarce environments: Challenges and
  opportunities.
\newblock {\em Frontiers in Environmental Science}, 7, 2019.

\bibitem{schuhmann2022laion}
Christoph Schuhmann, Romain Beaumont, Richard Vencu, Cade Gordon, Ross
  Wightman, Mehdi Cherti, Theo Coombes, Aarush Katta, Clayton Mullis, Mitchell
  Wortsman, et~al.
\newblock Laion-5b: An open large-scale dataset for training next generation
  image-text models.
\newblock {\em arXiv preprint arXiv:2210.08402}, 2022.

\bibitem{schuhmann2021laion}
Christoph Schuhmann, Richard Vencu, Romain Beaumont, Robert Kaczmarczyk,
  Clayton Mullis, Aarush Katta, Theo Coombes, Jenia Jitsev, and Aran
  Komatsuzaki.
\newblock Laion-400m: Open dataset of clip-filtered 400 million image-text
  pairs.
\newblock {\em arXiv preprint arXiv:2111.02114}, 2021.

\bibitem{sharma2018conceptual}
Piyush Sharma, Nan Ding, Sebastian Goodman, and Radu Soricut.
\newblock Conceptual captions: A cleaned, hypernymed, image alt-text dataset
  for automatic image captioning.
\newblock In {\em Proceedings of the 56th Annual Meeting of the Association for
  Computational Linguistics (Volume 1: Long Papers)}, pages 2556--2565, 2018.

\bibitem{srivastava2022beyond}
Aarohi Srivastava, Abhinav Rastogi, Abhishek Rao, Abu Awal~Md Shoeb, Abubakar
  Abid, Adam Fisch, Adam~R Brown, Adam Santoro, Aditya Gupta, Adri{\`a}
  Garriga-Alonso, et~al.
\newblock Beyond the imitation game: Quantifying and extrapolating the
  capabilities of language models.
\newblock {\em arXiv preprint arXiv:2206.04615}, 2022.

\bibitem{sumbul2019bigearthnet}
Gencer Sumbul, Marcela Charfuelan, Beg{\"u}m Demir, and Volker Markl.
\newblock Bigearthnet: A large-scale benchmark archive for remote sensing image
  understanding.
\newblock In {\em IGARSS 2019-2019 IEEE International Geoscience and Remote
  Sensing Symposium}, pages 5901--5904. IEEE, 2019.

\bibitem{thomee2016yfcc100m}
Bart Thomee, David~A Shamma, Gerald Friedland, Benjamin Elizalde, Karl Ni,
  Douglas Poland, Damian Borth, and Li-Jia Li.
\newblock Yfcc100m: The new data in multimedia research.
\newblock {\em Communications of the ACM}, 59(2):64--73, 2016.

\bibitem{GID2020}
Xin-Yi Tong, Gui-Song Xia, Qikai Lu, Huanfeng Shen, Shengyang Li, Shucheng You,
  and Liangpei Zhang.
\newblock Land-cover classification with high-resolution remote sensing images
  using transferable deep models.
\newblock {\em Remote Sensing of Environment}, 237:111322, 2020.

\bibitem{triantafillou2019meta}
Eleni Triantafillou, Tyler Zhu, Vincent Dumoulin, Pascal Lamblin, Utku Evci,
  Kelvin Xu, Ross Goroshin, Carles Gelada, Kevin Swersky, Pierre-Antoine
  Manzagol, et~al.
\newblock Meta-dataset: A dataset of datasets for learning to learn from few
  examples.
\newblock {\em arXiv preprint arXiv:1903.03096}, 2019.

\bibitem{wagner2018geographic}
K Wagner.
\newblock Geographic information systems and glacial environments.
\newblock In {\em Past Glacial Environments}, pages 503--536. Elsevier, 2018.

\bibitem{wang2019superglue}
Alex Wang, Yada Pruksachatkun, Nikita Nangia, Amanpreet Singh, Julian Michael,
  Felix Hill, Omer Levy, and Samuel Bowman.
\newblock Superglue: A stickier benchmark for general-purpose language
  understanding systems.
\newblock {\em Advances in neural information processing systems}, 32, 2019.

\bibitem{wang2018glue}
Alex Wang, Amanpreet Singh, Julian Michael, Felix Hill, Omer Levy, and Samuel~R
  Bowman.
\newblock Glue: A multi-task benchmark and analysis platform for natural
  language understanding.
\newblock {\em arXiv preprint arXiv:1804.07461}, 2018.

\bibitem{wang2018scene}
Qi Wang, Shaoteng Liu, Jocelyn Chanussot, and Xuelong Li.
\newblock Scene classification with recurrent attention of vhr remote sensing
  images.
\newblock {\em IEEE Transactions on Geoscience and Remote Sensing},
  57(2):1155--1167, 2018.

\bibitem{wolf-etal-2020-transformers}
Thomas Wolf, Lysandre Debut, Victor Sanh, Julien Chaumond, Clement Delangue,
  Anthony Moi, Pierric Cistac, Tim Rault, Rémi Louf, Morgan Funtowicz, Joe
  Davison, Sam Shleifer, Patrick von Platen, Clara Ma, Yacine Jernite, Julien
  Plu, Canwen Xu, Teven~Le Scao, Sylvain Gugger, Mariama Drame, Quentin Lhoest,
  and Alexander~M. Rush.
\newblock Transformers: State-of-the-art natural language processing.
\newblock In {\em Proceedings of the 2020 Conference on Empirical Methods in
  Natural Language Processing: System Demonstrations}, pages 38--45, Online,
  Oct. 2020. Association for Computational Linguistics.

\bibitem{xia2017aid}
Gui-Song Xia, Jingwen Hu, Fan Hu, Baoguang Shi, Xiang Bai, Yanfei Zhong,
  Liangpei Zhang, and Xiaoqiang Lu.
\newblock Aid: A benchmark data set for performance evaluation of aerial scene
  classification.
\newblock {\em IEEE Transactions on Geoscience and Remote Sensing},
  55(7):3965--3981, 2017.

\bibitem{Xia2010WHURS19}
Gui-Song Xia, Wen Yang, Julie Delon, Yann Gousseau, Hong Sun, and Henri
  MaÎtre.
\newblock Structural high-resolution satellite image indexing.
\newblock {\em Symposium: 100 Years ISPRS - Advancing Remote Sensing Science},
  2010.

\bibitem{xiao2017high}
Zhifeng Xiao, Yang Long, Deren Li, Chunshan Wei, Gefu Tang, and Junyi Liu.
\newblock High-resolution remote sensing image retrieval based on cnns from a
  dimensional perspective.
\newblock {\em Remote Sensing}, 9(7):725, 2017.

\bibitem{earthnets4eo}
Zhitong Xiong, Fahong Zhang, Yi Wang, Yilei Shi, and Xiao~Xiang Zhu.
\newblock Earthnets: Empowering ai in earth observation.
\newblock {\em arXiv:2210.04936}, 2022.

\bibitem{yang2010bag}
Yi Yang and Shawn Newsam.
\newblock Bag-of-visual-words and spatial extensions for land-use
  classification.
\newblock In {\em Proceedings of the 18th SIGSPATIAL international conference
  on advances in geographic information systems}, pages 270--279, 2010.

\bibitem{yao2021filip}
Lewei Yao, Runhui Huang, Lu Hou, Guansong Lu, Minzhe Niu, Hang Xu, Xiaodan
  Liang, Zhenguo Li, Xin Jiang, and Chunjing Xu.
\newblock Filip: fine-grained interactive language-image pre-training.
\newblock {\em arXiv preprint arXiv:2111.07783}, 2021.

\bibitem{zhai2019large}
Xiaohua Zhai, Joan Puigcerver, Alexander Kolesnikov, Pierre Ruyssen, Carlos
  Riquelme, Mario Lucic, Josip Djolonga, Andre~Susano Pinto, Maxim Neumann,
  Alexey Dosovitskiy, et~al.
\newblock A large-scale study of representation learning with the visual task
  adaptation benchmark.
\newblock {\em arXiv preprint arXiv:1910.04867}, 2019.

\bibitem{zhao2015dirichlet}
Bei Zhao, Yanfei Zhong, Gui-Song Xia, and Liangpei Zhang.
\newblock Dirichlet-derived multiple topic scene classification model for high
  spatial resolution remote sensing imagery.
\newblock {\em IEEE Transactions on Geoscience and Remote Sensing},
  54(4):2108--2123, 2015.

\bibitem{zhao2016fisher}
Bei Zhao, Yanfei Zhong, Liangpei Zhang, and Bo Huang.
\newblock The fisher kernel coding framework for high spatial resolution scene
  classification.
\newblock {\em Remote Sensing}, 8(2):157, 2016.

\bibitem{zhao2016feature}
Lijun Zhao, Ping Tang, and Lianzhi Huo.
\newblock Feature significance-based multibag-of-visual-words model for remote
  sensing image scene classification.
\newblock {\em Journal of Applied Remote Sensing}, 10(3):035004--035004, 2016.

\bibitem{zhou2018patternnet}
Weixun Zhou, Shawn Newsam, Congmin Li, and Zhenfeng Shao.
\newblock Patternnet: A benchmark dataset for performance evaluation of remote
  sensing image retrieval.
\newblock {\em ISPRS journal of photogrammetry and remote sensing},
  145:197--209, 2018.

\bibitem{Zhou2021NaSCTG2}
Zhuang Zhou, Shengyang Li, Wei Wu, Weilong Guo, Xuan Li, and Guisong
  Xiaand~Zifei Zhao.
\newblock Nasc-tg2: Natural scene classification with tiangong-2 remotely
  sensed imagery.
\newblock {\em IEEE Journal of Selected Topics in Applied Earth Observations
  and Remote Sensing}, 14:3228--3242, 2021.

\bibitem{zhu2016bag}
Qiqi Zhu, Yanfei Zhong, Bei Zhao, Gui-Song Xia, and Liangpei Zhang.
\newblock Bag-of-visual-words scene classifier with local and global features
  for high spatial resolution remote sensing imagery.
\newblock {\em IEEE Geoscience and Remote Sensing Letters}, 13(6):747--751,
  2016.

\bibitem{zhu2019so2sat}
Xiao~Xiang Zhu, Jingliang Hu, Chunping Qiu, Yilei Shi, Jian Kang, Lichao Mou,
  Hossein Bagheri, Matthias H{\"a}berle, Yuansheng Hua, Rong Huang, et~al.
\newblock So2sat lcz42: A benchmark dataset for global local climate zones
  classification.
\newblock {\em arXiv preprint arXiv:1912.12171}, 2019.

\bibitem{7272047}
Qin Zou, Lihao Ni, Tong Zhang, and Qian Wang.
\newblock Deep learning based feature selection for remote sensing scene
  classification.
\newblock {\em IEEE Geoscience and Remote Sensing Letters}, 12(11):2321--2325,
  2015.

\end{thebibliography}
}
\end{multicols}
\clearpage
\appendix
\section{Appendix}
\label{section:appendix}

We structure this Appendix to our main paper submission as follows. First, we outline our reproducibility statement, detailing how other researchers can replicate our findings (Sec. \ref{section:reproducibility}). Following this, we outline the structure, functionality and benefits of our public leaderboard (Sec. \ref{section:website}). Then, we provide a more detailed overview of the constituent datasets of SATIN, including: licencing details, split information, class names, and more example imagery (Sec. \ref{section:datasets}). Next, we present our results in more detail, with \textit{per dataset} scores; we also include a more in-depth analysis of the performance of the CLIP model across SATIN (Sec. \ref{section:results}). We follow this by providing pseudocode for our inference pipeline as well as a table containing links to the implementation scripts and checkpoints for the models that we evaluate, and a table for the prompts we used for each dataset (Sec. \ref{section:methodology}). Finally, we give a comparison of the compute requirements for evaluating SATIN for each of the 8 methods (Sec. \ref{section:compute_requirements}).

\vspace{2cm}

\subsection{Reproducibility}
\label{section:reproducibility}
We curate SATIN to be a useful and easy-to-use addition for the remote sensing and wider computer vision community.
As such, we aim for our work to be entirely reproducible and release SATIN via both HuggingFace and Zenodo.
We make use of publicly available model checkpoints and ensure attribution to both the checkpoint and the corresponding methodology on which we based our implementation of that particular model.
We outline our methodology, including our inference pipeline, prompting strategy and model selection, to allow our results and wider findings to be reproduced.
Furthermore, we will release our code upon publication.

\vspace{2cm}

\subsection{Public Leaderboard}
\label{section:website}
When released, our website will be hosted using GitHub Pages featuring a design that draws inspiration from popular metadataset benchmark websites such as \href{https://gluebenchmark.com/}{GLUE}, \href{https://super.gluebenchmark.com/}{SuperGLUE} and \href{https://rajpurkar.github.io/SQuAD-explorer/}{SQuAD}. The site will include: (i) a SATIN public leaderboard; (ii) code to download and use SATIN; (iii) a more detailed overview of the constituent datasets and tasks; and (iv) links to our paper, blog post, and other useful material. There are numerous benefits to hosting a website with a leaderboard: (1) \textbf{Progress tracking}---Hosting a public leaderboard conglomerates the performances of other models into a single location, allowing users to directly compare the performance of their models with existing works, without having to scour through the literature. The leaderboard is also dynamic and not restricted to the information available at the time of publication, meaning it can serve as a useful tool for tracking the progress of research in this domain. (2) \textbf{Ease of use}---The website will provide a one-stop shop for SATIN, giving easy access to the leaderboard, code, datasets and other useful information. (3) \textbf{Living benchmark}---A further benefit of the dynamic nature of a website is that it allows SATIN to be a living benchmark that is incrementally updated with expansions and improvements, and can be modified in response to feedback from the community. The combination of these three factors enables SATIN to facilitate VL research in the remote sensing domain and drive progress forwards.

\clearpage
\subsection{Datasets}
\label{section:datasets}
The following section outlines additional details of the constituent datasets in SATIN. 

\subsubsection{Licencing Details}
\label{appendix:licencing_details}

As mentioned in the paper, we only include datasets without licence or usage restrictions that prevent them from being used for academic research. There is a wide variety of different licences attributed to each dataset and hence their downstream usage differs slightly -- see Table \ref{table:licences} for details.
Note that we do not give SATIN an overall licence as the constituent datasets hold different licences. Instead, we retain the individual licences and usage restrictions specified for each dataset.

\begin{table}[!h]
    \small
    \centering
    \begin{tabular}{l|p{5cm} p{5.5cm}}
    \hline
    \textbf{Dataset} & \textbf{Licence/Usage} & \textbf{Source of Licence/Usage Information}\\
    \hline
    SAT-4 & \multirow{2}{*}{Public Domain} & \multirow{2}{*}{\begin{tabular}{@{}l@{}}Obtained via correspondence with author(s) \\ + \href{https://eos.com/find-satellite/naip/}{EOS}\end{tabular}} \\
    SAT-6 \\
    \hline
    NaSC-TG2 & CC BY-NC-SA 4.0 & Obtained via correspondence with author(s) \\
    \hline
    WHU-RS19 & CC BY 4.0 & \href{https://figshare.com/articles/dataset/Untitled_Item/6086159}{Figshare} \\
    \hline
    RSSCN7 & \textit{For research/academic purposes} & Obtained via correspondence with author(s) \\
    \hline
    RSC11 & \textit{Free usage without licence} & Obtained via correspondence with author(s) \\
    \hline
    SIRI-WHU & CC BY-NC-ND & Obtained via correspondence with author(s) \\
    \hline
    EuroSAT & MIT & Obtained via correspondence with author(s) \\
    \hline
    NWPU-RESISC45 & CC-BY-SA & Obtained via correspondence with author(s) \\
    \hline
    PatternNet & \textit{For research purposes} & Obtained via correspondence with author(s) \\
    \hline
    RSD46-WHU & \textit{For education/research/commercial} & \href{https://github.com/RSIA-LIESMARS-WHU/RSD46-WHU}{GitHub} \\
    \hline
    GID & Public Domain & Obtained via correspondence with author(s) \\
    \hline
    CLRS & \textit{For academic purposes} & Obtained via correspondence with author(s) \\
    \hline
    Optimal-31 & \textit{No licence, cite paper} & Obtained via correspondence with author(s) \\
    \hline
    RSI-CB256 & \textit{For academic purposes} & Obtained via correspondence with author(s) \\
    \hline
    Million-AID & CC BY-NC-ND 4.0 & \href{https://competitions.codalab.org/competitions/35974#learn_the_details-terms-and-conditions}{CodaLab} \\
    \hline
    UCM Land Use & Public Domain$^*$ & Obtained via correspondence with author(s) + \href{https://www.usgs.gov/faqs/what-are-terms-uselicensing-map-services-and-data-national-map}{USGS} \\
    \hline
    MLRSNet & CC BY 4.0 & \href{https://data.mendeley.com/datasets/7j9bv9vwsx/3}{Mendeley Data} \\
    \hline
    AID & CC0 1.0 Universal & \href{https://www.kaggle.com/datasets/jiayuanchengala/aid-scene-classification-datasets}{Kaggle} \\
    \hline
    MultiScene & MIT & Obtained via correspondence with author(s) \\
    \hline
    Airbus Wind Turbines Patches (AWTP) & CC BY-NC-SA 4.0 & \href{https://www.kaggle.com/datasets/airbusgeo/airbus-wind-turbines-patches}{Kaggle} \\
    \hline
    USTC\_SmokeRS & \textit{For research/education} & Obtained via correspondence with author(s) \\
    \hline
    Canadian Cropland & Montreal Data Licence & \href{https://github.com/bioinfoUQAM/Canadian-cropland-dataset/blob/main/DATA_LICENSE}{GitHub} \\
    \hline
    Ships In Satellite Imagery & CC BY-SA 4.0 & \href{https://www.kaggle.com/datasets/rhammell/ships-in-satellite-imagery}{Kaggle} \\
    \hline
    Post Hurricane & CC BY 4.0 & \href{https://ieee-dataport.org/open-access/detecting-damaged-buildings-post-hurricane-satellite-imagery-based-customized}{IEEE Dataport} \\
    \hline
    BC Scenes & CC BY-NC & Obtained via correspondence with author(s) \\
    \hline
    BCS Scenes & CC BY-NC & Obtained via correspondence with author(s) \\
    \hline
    \end{tabular}
    \vspace{0.25cm}
    \caption{\textbf{SATIN constituent dataset licencing and usage information}. $^*$ No restrictions except a request to include the following in derived products and data: ``Map services and data available from U.S. Geological Survey, National Geospatial Program.''}
    \label{table:licences}
\end{table}

\clearpage

\subsubsection{Split Information}
\label{appendix:subsets}

We use the majority of datasets in SATIN in their entirety, however, for some we use a subset -- see Table \ref{table:splits} for descriptions of the splits used where appropriate.

\begin{table}[!h]
    \small
    \centering
    \begin{tabular}{l|l}
    \hline
    \textbf{Dataset} & \textbf{Split used in SATIN} \\ 
    \hline
    SAT-4 & \textit{Test} \\
    SAT-6 & \textit{Test} \\
    SIRI-WHU & \textit{Google Image Dataset} \\
    RSD46-WHU & \textit{Validation} \\
    GID & \textit{Fine Classification, Train}\\
    MillionAID & \textit{Train} \\
    MultiScene & \textit{Clean} \\
    Airbus Wind Turbine Patches & \textit{Validation}\\
    Post Hurricane & \textit{Train\_Another} \\
    Canadian Cropland & \textit{2017} \\    
    \hline
    \end{tabular}
    \vspace{0.25cm}
    \caption{\textbf{SATIN dataset splits}. For datasets not included in this table, all splits are used.} 
    \label{table:splits}
\end{table}

\clearpage
\subsubsection{Example Imagery}
\begin{figure*}[!ht]
\centering
\scalebox{0.9}{
\begin{tabular}{ccp{0.02cm}cc}
  \includegraphics[width=\ImWidth]{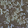} &  \includegraphics[width=\ImWidth]{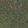} 
  & & 
  \includegraphics[width=\ImWidth]{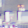} & \includegraphics[width=\ImWidth]{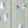} \\
  trees & grassland
  & & 
  building & road \\
  \multicolumn{2}{c}{\textbf{SAT-4}} 
  & & 
  \multicolumn{2}{c}{\textbf{SAT-6}} \\

  \multicolumn{5}{c}{\rule{0pt}{5pt}} \\

  \includegraphics[width=\ImWidth]{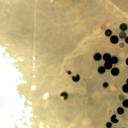} &  \includegraphics[width=\ImWidth]{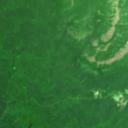} 
  & & 
  \includegraphics[width=\ImWidth]{figures/example_images/WHU-19_footballfield_1.jpg} & \includegraphics[width=\ImWidth]{figures/example_images/WHU-19_viaduct_07.jpg} \\
  circular farmland & forest
  & &
  football field & viaduct \\
\multicolumn{2}{c}{\textbf{NaSC-TG2}} 
& & \multicolumn{2}{c}{\textbf{WHU-RS19}} \\

  \multicolumn{5}{c}{\rule{0pt}{\RowHeight}} \\

  \includegraphics[width=\ImWidth]{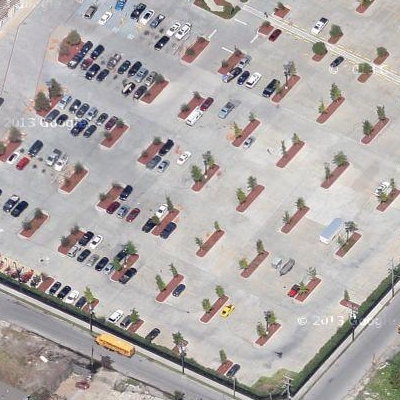} &  \includegraphics[width=\ImWidth]{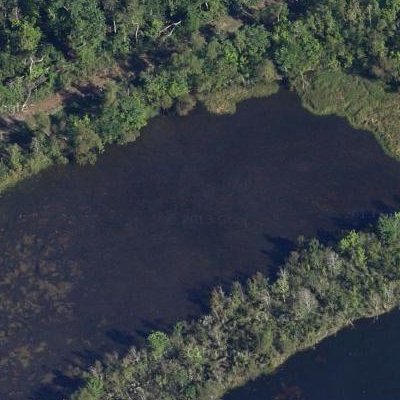} 
  & & 
  \includegraphics[width=\ImWidth]{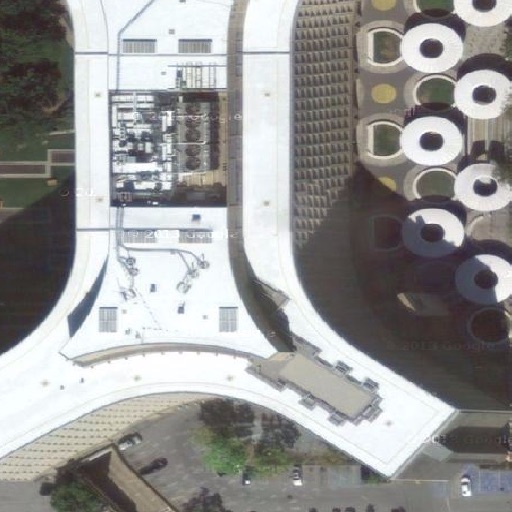} & \includegraphics[width=\ImWidth]{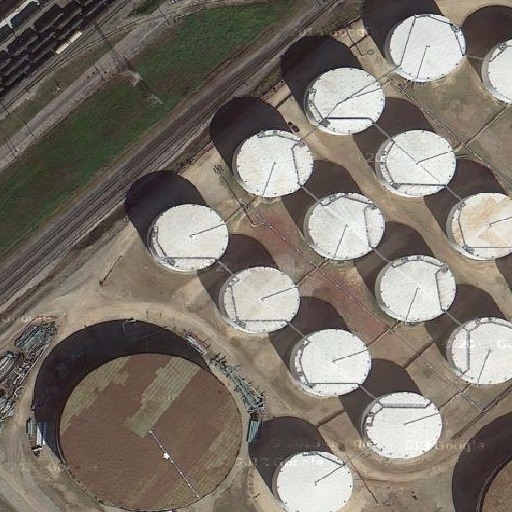} \\
  parking & river or lake
  & &
  high buildings & storage tanks \\
\multicolumn{2}{c}{\textbf{RSSCN7}} 
& & \multicolumn{2}{c}{\textbf{RSC11}} \\

  \multicolumn{5}{c}{\rule{0pt}{\RowHeight}} \\

  \includegraphics[width=\ImWidth]{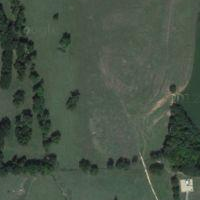} &  \includegraphics[width=\ImWidth]{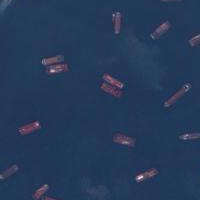} 
  & & 
  \includegraphics[width=\ImWidth]{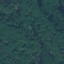} & \includegraphics[width=\ImWidth]{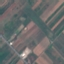} \\
  meadow & water
  & &
  forest & permanent crop \\
\multicolumn{2}{c}{\textbf{SIRI-WHU}} 
& & \multicolumn{2}{c}{\textbf{EuroSAT}} \\


\end{tabular}}
\vspace{-0.1cm}
\caption{\textbf{Example SATIN imagery and ground truth labels} for \textit{Task 1: Land Cover} (SAT-4 -- NaSC-TG2) and \textit{Task 2: Land Use} (WHU-RS19 -- EuroSAT).
}
\label{fig:example_imagery_t1_2}
\end{figure*}

\begin{figure*}[p]
\centering
\scalebox{0.9}{
\begin{tabular}{ccp{0.02cm}cc}

  \includegraphics[width=\ImWidth]{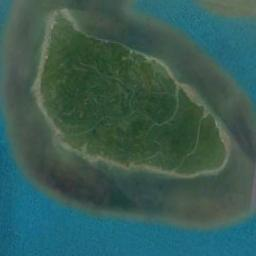} &  \includegraphics[width=\ImWidth]{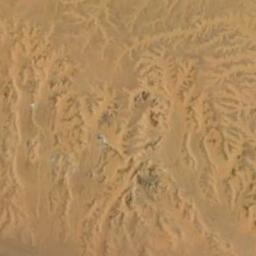} 
  & & 
  \includegraphics[width=\ImWidth]{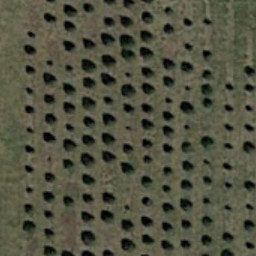} & \includegraphics[width=\ImWidth]{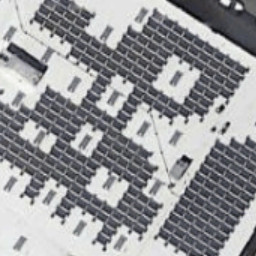} \\
  island & desert
  & &
  christmas tree farm & solar panel \\
\multicolumn{2}{c}{\textbf{NWPU-RESISC45}} 
& & \multicolumn{2}{c}{\textbf{PatternNet}} \\

 \multicolumn{5}{c}{\rule{0pt}{\RowHeight}} \\

  \includegraphics[width=\ImWidth]{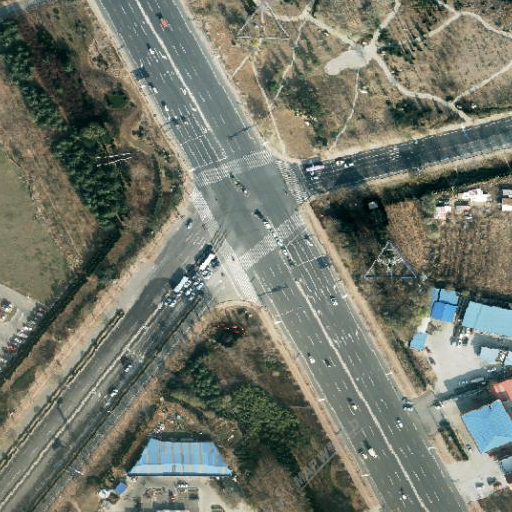} &  \includegraphics[width=\ImWidth]{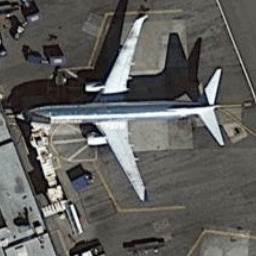} 
  & & 
  \includegraphics[width=\ImWidth]{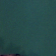} & \includegraphics[width=\ImWidth]{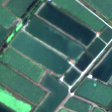} \\
  crossroads & airplane
  & &
  lake & paddy field \\
\multicolumn{2}{c}{\textbf{RSD46-WHU}} 
& & \multicolumn{2}{c}{\textbf{GID}} \\

 \multicolumn{5}{c}{\rule{0pt}{\RowHeight}} \\

  \includegraphics[width=\ImWidth]{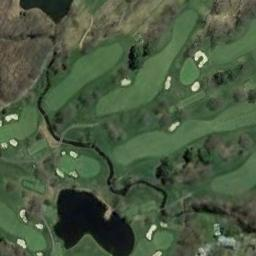} &  \includegraphics[width=\ImWidth]{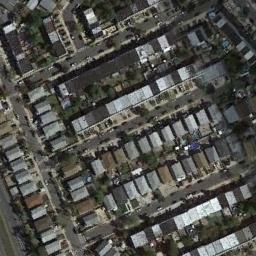} 
  & & 
  \includegraphics[width=\ImWidth]{figures/example_images/CLRS_10384_port.png} & \includegraphics[width=\ImWidth]{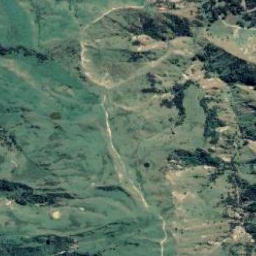} \\
  golf course & dense residential
  & &
  port & mountain \\
\multicolumn{2}{c}{\textbf{Optimal-31}} 
& & \multicolumn{2}{c}{\textbf{CLRS}} \\

\end{tabular}}
\label{fig:example_imagery_t2}
\caption{\textbf{Example SATIN imagery and ground truth labels} for the remaining \textit{Task 2: Land Use} datasets.}
\end{figure*}

\begin{figure*}[!ht]
\centering
\scalebox{0.9}{
\begin{tabular}{ccp{0.02cm}cc}

  \includegraphics[width=\ImWidth]{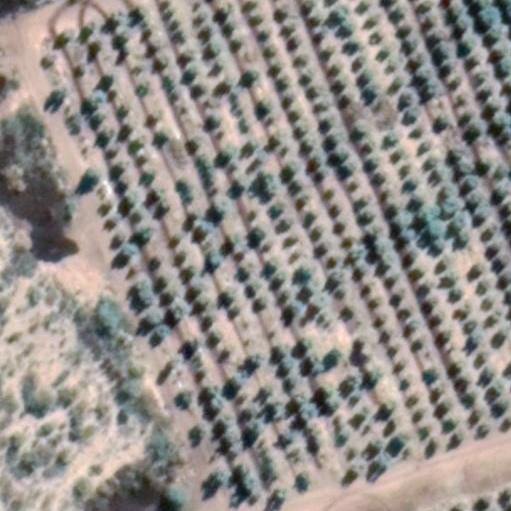} &  \includegraphics[width=\ImWidth]{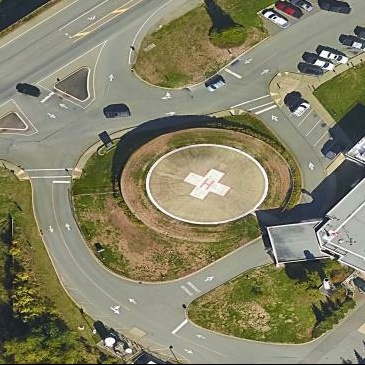} 
  & & 
  \includegraphics[width=\ImWidth]{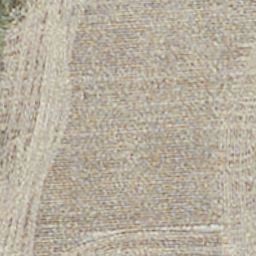} & \includegraphics[width=\ImWidth]{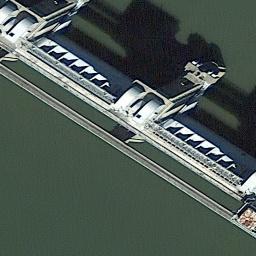} \\
  \begin{tabular}{@{}c@{}}agricultural land $\rightarrow$ \\ woodland $\rightarrow$ orchard \end{tabular}  & \begin{tabular}{@{}c@{}}transportation land $\rightarrow$ \\ airport area $\rightarrow$ helipad \end{tabular}
  & &
  \begin{tabular}{@{}c@{}}cultivated land $\rightarrow$ \\ dry farm \end{tabular} & \begin{tabular}{@{}c@{}}water area $\rightarrow$ \\ dam \end{tabular} \\
\multicolumn{2}{c}{\textbf{Million-AID}} 
& & \multicolumn{2}{c}{\textbf{RSI-CB256}} \\

 \multicolumn{5}{c}{\rule{0pt}{\RowHeight}} \\

  \includegraphics[width=\ImWidth]{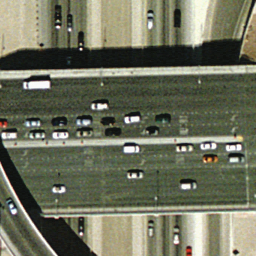} &  \includegraphics[width=\ImWidth]{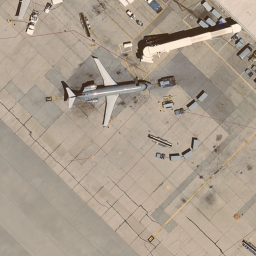} 
  & & 
  \includegraphics[width=\ImWidth]{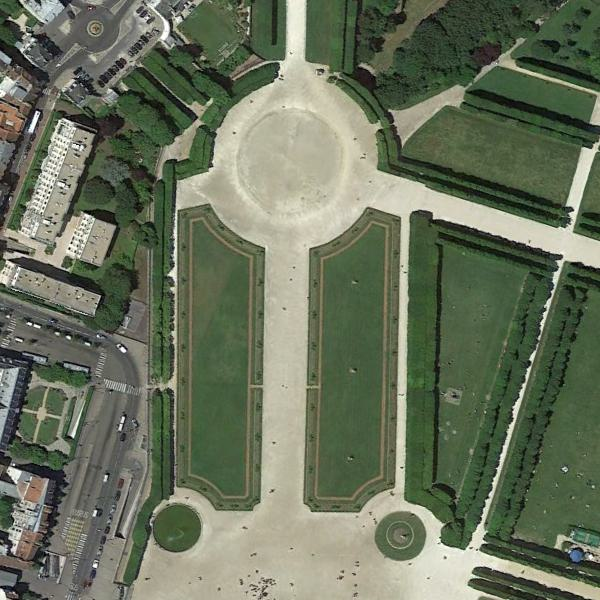} & \includegraphics[width=\ImWidth]{figures/example_images/AID_MultiLabel_317_sand_sea.png} \\
  \begin{tabular}{@{}c@{}}bare soil, cars \\ pavement \end{tabular} & \begin{tabular}{@{}c@{}}airplane, cars \\ pavement \end{tabular}
  & &
  \begin{tabular}{@{}c@{}} buildings, cars \\ grass, pavement, trees \end{tabular} & sand, sea \\
\multicolumn{2}{c}{\textbf{UCM Land Use}} 
& & \multicolumn{2}{c}{\textbf{AID}} \\

 \multicolumn{5}{c}{\rule{0pt}{\RowHeight}} \\

  \includegraphics[width=\ImWidth]{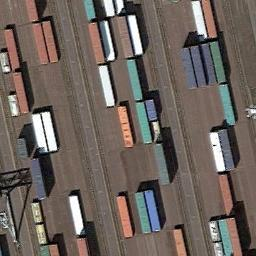} &  \includegraphics[width=\ImWidth]{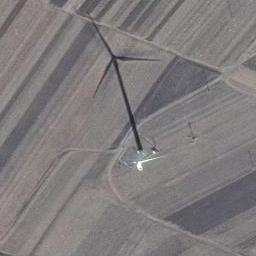} 
  & & 
  \includegraphics[width=\ImWidth]{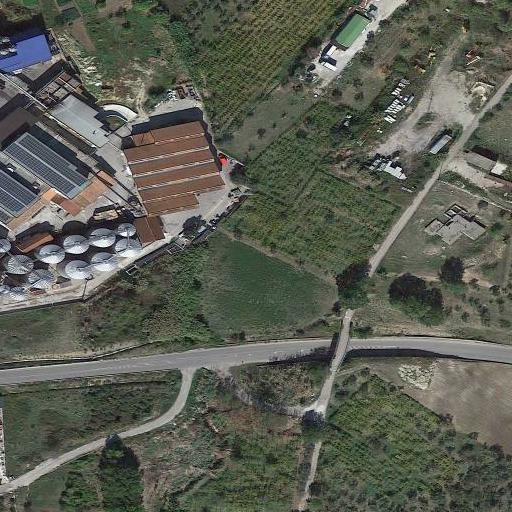} & \includegraphics[width=\ImWidth]{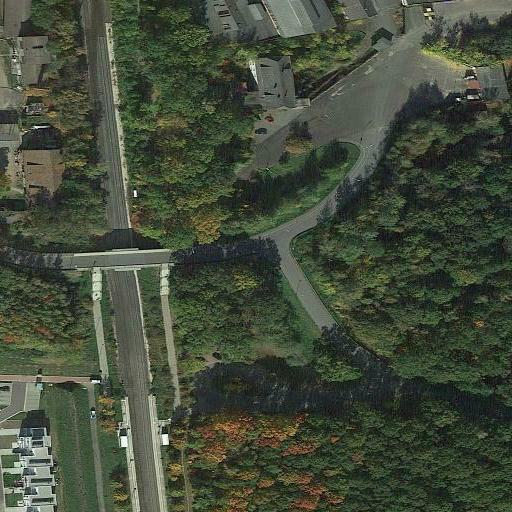} \\
  containers, pavement & \begin{tabular}{@{}c@{}} field, trail \\ wind turbine \end{tabular}
  & &
  \begin{tabular}{@{}c@{}} bridge, orchard \\ solar panel, storage tanks, works \end{tabular} & \begin{tabular}{@{}c@{}} bridge, commercial, woodland \\ parking lot, railway, residential \end{tabular} \\
\multicolumn{2}{c}{\textbf{MLRSNet}} 
& & \multicolumn{2}{c}{\textbf{MultiScene}} \\

\end{tabular}
}
\label{fig:example_imagery_t34}
\caption{\textbf{Example SATIN imagery and ground truth labels} for \textit{Task 3: Hierarchical Land Use} (Million-AID -- RSI-CB256) and \textit{Task 4: Complex Scenes} (UCM Land Use -- MultiScene).}

\end{figure*}

\begin{figure*}[!ht]
\centering
\scalebox{0.9}{
\begin{tabular}{ccp{0.02cm}cc}

\includegraphics[width=\ImWidth]{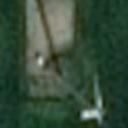} &  \includegraphics[width=\ImWidth]{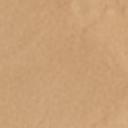} 
  & & 
  \includegraphics[width=\ImWidth]{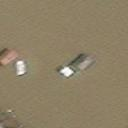} & \includegraphics[width=\ImWidth]{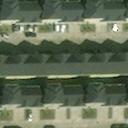} \\
  wind turbine & no wind turbine
  & &
  damaged/flooded buildings & undamaged buildings \\
\multicolumn{2}{c}{\textbf{Airbus Wind Turbine Patches (AWTP)}} 
& & \multicolumn{2}{c}{\textbf{Post Hurricane}} \\

 \multicolumn{5}{c}{\rule{0pt}{\RowHeight}} \\

  \includegraphics[width=\ImWidth]{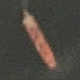} &  \includegraphics[width=\ImWidth]{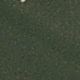} 
  & & 
  \includegraphics[width=\ImWidth]{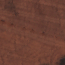} & \includegraphics[width=\ImWidth]{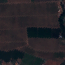} \\
  an entire ship & no ship or part of a ship
  & &
  barley & mixedwood \\
\multicolumn{2}{c}{\textbf{Ships in Satellite Imagery (SISI)}} 
& & \multicolumn{2}{c}{\textbf{Canadian Cropland}} \\

 \multicolumn{5}{c}{\rule{0pt}{\RowHeight}} \\

    & \includegraphics[width=\ImWidth]{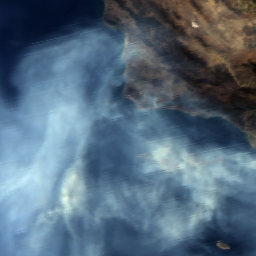} & 
   \includegraphics[width=\ImWidth]{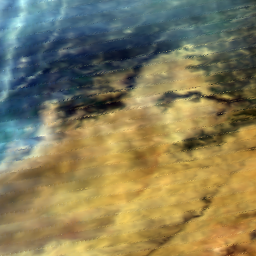} & \\
  & smoke &  & \hspace{-9mm} haze &  \\
  & \multicolumn{3}{c}{\textbf{USTC-SmokeRS}} & \\

 \multicolumn{5}{c}{\rule{0pt}{\RowHeight}} \\
  
  \includegraphics[width=\ImWidth]{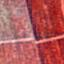} & \includegraphics[width=\ImWidth]{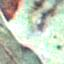}
  && 
  \includegraphics[width=\ImWidth]{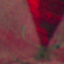} &  \includegraphics[width=\ImWidth]{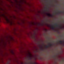} \\
  coffee & no coffee
  & &
  agriculture & arboreal vegetation \\
\multicolumn{2}{c}{\textbf{Brazilian Coffee Scenes (BC Scenes)}} 
& & \multicolumn{2}{c}{\textbf{Brazilian Cerrado-Savanna Scenes (BCS Scenes)}} \\

 \multicolumn{5}{c}{\rule{0pt}{\RowHeight}} \\

\end{tabular}}
\caption{\textbf{Example SATIN imagery and ground truth labels} for \textit{Task 5: Rare Scenes} (AWTP -- USTC-SmokeRS) and \textit{Task 6: False-Colour Scenes} (BC Scenes -- BCS Scenes).}
\label{fig:example_imagery_t56}
\end{figure*}

Figures \ref{fig:example_imagery_t1_2}--\ref{fig:example_imagery_t56} contain example imagery and ground truth labels for each of the constituent datasets in SATIN, ordered according to task. These examples illustrate the breadth of diversity in the SATIN benchmark.

\vspace{-0.25cm}


\clearpage


\subsection{Results}
\label{section:results}
This section presents the results for the models evaluated in the main paper in more detail. In Table \ref{table:8-method-detailed-results} we include per-dataset accuracy results for each of the 8 models outlined in Table 3 of the main paper. In Table \ref{table:remaining_results} we include the per-dataset accuracy results for the remaining models evaluated as part of the broad profile of VL models we benchmarked on SATIN. Tables \ref{table:8-method-detailed-results} and \ref{table:remaining_results} combined include the per dataset results for all the models presented in Figure 4 of the main paper. We then provide a deeper analysis of the performance on SATIN of the highest capacity CLIP model (ViT-L/14@336px, 400M) -- see Figures \ref{fig:bars}, \ref{fig:clip_confusionmatrices}.

\subsubsection{Per-Dataset Results}
\begin{table}[!ht]
    \small
    \centering
    \begin{tabular}{ l | c c c c c c c c }
    \hline
    \multicolumn{1}{c}{\textbf{Dataset}} & \multicolumn{8}{c}{\textbf{Zero-Shot Classification Accuracy}} \\
    \hline
    \textbf{} & \textbf{CyCLIP} & \textbf{ALBEF} & \textbf{SLIP} & \textbf{DeCLIP} & \textbf{BLIP} & \textbf{BLIP2} & \textbf{CLIP} & \textbf{OpenCLIP} \\
    \hline
SAT-4 & 0.19 & 0.38 & 0.21 & 0.41 & 0.33 & 0.44 & 0.45 & \textbf{0.54} \\
SAT-6 & 0.16 & 0.29 & 0.43 & 0.45 & 0.39 & 0.44 & \textbf{0.61} & 0.45 \\
NaSC-TG2 & 0.25 & 0.35 & 0.28 & 0.44 & 0.40 & 0.54 & 0.54 & \textbf{0.58} \\
    \hline
WHU-RS19 & 0.33 & 0.61 & 0.26 & 0.70 & 0.77 & 0.78 & 0.84 & \textbf{0.88} \\
RSSCN7 & 0.46 & 0.52 & 0.45 & 0.59 & 0.60 & 0.56 & 0.63 & \textbf{0.66} \\
RSC11 & 0.39 & 0.50 & 0.31 & 0.52 & 0.62 & 0.51 & 0.62 & \textbf{0.63} \\
SIRI-WHU & 0.19 & 0.40 & 0.23 & 0.38 & 0.47 & \textbf{0.63} & 0.59 & 0.56 \\
EuroSAT & 0.14 & 0.26 & 0.23 & 0.22 & 0.37 & \textbf{0.63} & 0.56 & 0.62 \\
NWPU-RESISC45 & 0.19 & 0.33 & 0.16 & 0.48 & 0.56 & 0.62 & 0.69 & \textbf{0.73} \\
PatternNet & 0.17 & 0.36 & 0.22 & 0.53 & 0.56 & 0.62 & 0.74 & \textbf{0.80} \\
RSD46-WHU & 0.08 & 0.20 & 0.10 & 0.20 & 0.29 & 0.37 & 0.36 & \textbf{0.44} \\
GID & 0.15 & 0.24 & 0.17 & 0.24 & 0.24 & 0.24 & 0.30 & \textbf{0.31} \\
CLRS & 0.28 & 0.39 & 0.22 & 0.52 & 0.61 & 0.62 & 0.61 & \textbf{0.68} \\
Optimal-31 & 0.27 & 0.49 & 0.23 & 0.58 & 0.69 & 0.71 & 0.77 & \textbf{0.84} \\
    \hline
RSI-CB256 & 0.35 & 0.36 & 0.20 & 0.40 & 0.46 & \textbf{0.49} & \textbf{0.49} & 0.46 \\
Million-AID & 0.20 & 0.30 & 0.18 & 0.35 & 0.44 & 0.51 & 0.53 & \textbf{0.54} \\
    \hline
UCM Land Use & 0.49 & 0.54 & 0.39 & 0.58 & \textbf{0.60} & 0.59 & 0.56 & 0.58 \\
MLRSNet & 0.30 & 0.29 & 0.17 & 0.32 & 0.38 & 0.38 & 0.37 & \textbf{0.39} \\
AID & 0.52 & 0.57 & 0.46 & 0.61 & \textbf{0.63} & 0.58 & 0.53 & 0.57 \\
MultiScene & 0.25 & 0.22 & 0.25 & 0.39 & 0.42 & \textbf{0.50} & \textbf{0.50} & 0.47 \\
    \hline
AWTP & 0.55 & 0.50 & 0.43 & 0.40 & 0.38 & 0.55 & \textbf{0.66} & 0.57 \\
USTC-SmokeRS & 0.16 & 0.24 & 0.39 & 0.34 & 0.46 & \textbf{0.55} & 0.50 & 0.53 \\
Canadian Cropland & 0.10 & 0.15 & 0.10 & \textbf{0.18} & 0.10 & \textbf{0.18} & 0.10 & 0.11 \\
SISI & 0.30 & \textbf{0.76} & 0.66 & 0.22 & 0.73 & 0.75 & 0.75 & 0.75 \\
Post Hurricane & 0.61 & 0.69 & 0.54 & 0.57 & 0.51 & \textbf{0.80} & 0.64 & 0.62 \\
    \hline
BC Scenes & \textbf{0.58} & 0.50 & 0.50 & 0.57 & 0.55 & 0.51 & 0.43 & 0.48 \\
BCS Scenes & 0.15 & 0.53 & 0.10 & \textbf{0.60} & 0.39 & 0.09 & 0.28 & 0.21 \\
\hline
\hline
\textbf{SATIN} & 0.25 & 0.38 & 0.26 & 0.41 & 0.45 & 0.48 & 0.51 & \textbf{0.52}\\
\hline
    \end{tabular}
    \vspace{0.25cm}
    \caption{\textbf{Per-dataset results for the main 8 models}. Best performance on each dataset and SATIN is in \textbf{bold}.
    Model details are given in Table 3 of the main paper.}
    \label{table:8-method-detailed-results}
\end{table}

\clearpage

\begin{table}
\begin{adjustbox}{angle=90}
\footnotesize
\centering
\setlength\tabcolsep{1.625pt}
\begin{tabular}[!t]{lll||lll|lllllllllll|ll|llll|lllll|ll}
\multicolumn{3}{c}{\textbf{Model Configuration}}                           & \multicolumn{27}{c}{\textbf{Zero-Shot Classification Accuracy}}  \\                                                   \hline                                                                                      
\multicolumn{3}{c}{} & \multicolumn{3}{c}{Task 1} & \multicolumn{11}{c}{Task 2} & \multicolumn{2}{c}{Task 3} & \multicolumn{4}{c}{Task 4} & \multicolumn{5}{c}{Task 5} & \multicolumn{2}{c}{Task 6}\\  

\rotatebox{\RotateAngle}{Method}   & \rotatebox{\RotateAngle}{Backbone}                & \rotatebox{\RotateAngle}{Pretraining} & \rotatebox{\RotateAngle}{SAT-4} & \rotatebox{\RotateAngle}{SAT-6} & \rotatebox{\RotateAngle}{NASC-TG2} & \rotatebox{\RotateAngle}{WHU-RS19} & \rotatebox{\RotateAngle}{RSSCN7} & \rotatebox{\RotateAngle}{RSC11} & \rotatebox{\RotateAngle}{SIRI-WHU} & \rotatebox{\RotateAngle}{EuroSAT} & \rotatebox{\RotateAngle}{NWPU-RESISC45} & \rotatebox{\RotateAngle}{PatternNet} & \rotatebox{\RotateAngle}{RSD46-WHU} & \rotatebox{\RotateAngle}{GID}  & \rotatebox{\RotateAngle}{CLRS} & \rotatebox{\RotateAngle}{Optimal-31} & \rotatebox{\RotateAngle}{Million-AID} & \rotatebox{\RotateAngle}{RSI-CB256} & \rotatebox{\RotateAngle}{UCM Land Use} & \rotatebox{\RotateAngle}{MLRSNet} & \rotatebox{\RotateAngle}{AID}  & \rotatebox{\RotateAngle}{MultiScene} & \rotatebox{\RotateAngle}{AWTP} & \rotatebox{\RotateAngle}{USTC-SmokeRS} & \rotatebox{\RotateAngle}{Canadian Cropland} & \rotatebox{\RotateAngle}{SISI} & \rotatebox{\RotateAngle}{Post Hurricane} & \rotatebox{\RotateAngle}{BC Scenes} & \rotatebox{\RotateAngle}{BCS Scenes} \\
\hline
CyCLIP   & RN50                    & 3M                    & 0.19  & 0.16  & 0.25     & 0.33     & 0.46   & 0.39  & 0.19     & 0.14    & 0.19          & 0.17       & 0.08      & 0.15 & 0.28 & 0.27       & 0.20        & 0.35      & 0.49         & 0.30    & 0.52 & 0.25       & 0.55 & 0.16         & 0.10              & 0.30 & 0.61           & 0.58      & 0.15       \\
CLIP     & RN50                    & 3M                    & 0.22  & 0.39  & 0.29     & 0.31     & 0.43   & 0.39  & 0.16     & 0.15    & 0.21          & 0.15       & 0.09      & 0.16 & 0.25 & 0.29       & 0.21        & 0.20      & 0.50         & 0.30    & 0.48 & 0.29       & 0.56 & 0.21         & 0.09              & 0.27 & 0.59           & 0.53      & 0.64       \\
CyCLIP   & RN50                    & 2.6M                  & 0.24  & 0.18  & 0.23     & 0.35     & 0.42   & 0.42  & 0.21     & 0.11    & 0.22          & 0.20       & 0.11      & 0.17 & 0.30 & 0.33       & 0.19        & 0.28      & 0.45         & 0.27    & 0.52 & 0.34       & 0.57 & 0.17         & 0.12              & 0.27 & 0.58           & 0.62      & 0.08       \\
CLIP     & RN50                    & 2.6M                  & 0.28  & 0.19  & 0.26     & 0.33     & 0.48   & 0.50  & 0.17     & 0.29    & 0.18          & 0.17       & 0.10      & 0.18 & 0.26 & 0.26       & 0.16        & 0.22      & 0.45         & 0.29    & 0.53 & 0.31       & 0.50 & 0.15         & 0.13              & 0.73 & 0.50           & 0.64      & 0.14       \\
CyCLIP   & RN50                    & 2M                    & 0.22  & 0.19  & 0.19     & 0.30     & 0.47   & 0.49  & 0.17     & 0.20    & 0.17          & 0.16       & 0.08      & 0.17 & 0.25 & 0.24       & 0.20        & 0.30      & 0.48         & 0.31    & 0.56 & 0.28       & 0.57 & 0.21         & 0.06              & 0.37 & 0.58           & 0.49      & 0.76       \\
CLIP     & RN50                    & 2M                    & 0.24  & 0.20  & 0.30     & 0.29     & 0.41   & 0.48  & 0.24     & 0.23    & 0.19          & 0.16       & 0.09      & 0.21 & 0.24 & 0.27       & 0.20        & 0.30      & 0.42         & 0.23    & 0.50 & 0.30       & 0.57 & 0.20         & 0.09              & 0.41 & 0.68           & 0.48      & 0.19       \\
CyCLIP   & RN50                    & 1M                    & 0.21  & 0.18  & 0.22     & 0.33     & 0.39   & 0.38  & 0.21     & 0.15    & 0.19          & 0.16       & 0.07      & 0.14 & 0.23 & 0.28       & 0.22        & 0.30      & 0.43         & 0.25    & 0.53 & 0.25       & 0.54 & 0.21         & 0.11              & 0.54 & 0.65           & 0.52      & 0.14       \\
CLIP     & RN50                    & 1M                    & 0.20  & 0.47  & 0.24     & 0.27     & 0.41   & 0.41  & 0.20     & 0.21    & 0.17          & 0.13       & 0.08      & 0.19 & 0.18 & 0.24       & 0.19        & 0.28      & 0.40         & 0.27    & 0.49 & 0.28       & 0.56 & 0.17         & 0.07              & 0.72 & 0.57           & 0.52      & 0.72       \\
CyCLIP   & RN50                    & 500K                  & 0.36  & 0.36  & 0.18     & 0.24     & 0.37   & 0.28  & 0.19     & 0.18    & 0.11          & 0.08       & 0.06      & 0.18 & 0.18 & 0.16       & 0.17        & 0.32      & 0.45         & 0.27    & 0.48 & 0.24       & 0.54 & 0.20         & 0.13              & 0.77 & 0.62           & 0.45      & 0.08       \\
CLIP     & RN50                    & 500K                  & 0.28  & 0.39  & 0.22     & 0.23     & 0.38   & 0.28  & 0.22     & 0.19    & 0.14          & 0.11       & 0.07      & 0.16 & 0.19 & 0.18       & 0.17        & 0.31      & 0.44         & 0.26    & 0.47 & 0.25       & 0.37 & 0.21         & 0.17              & 0.63 & 0.58           & 0.62      & 0.41       \\
C-CyCLIP & RN50                    & 2.6M                  & 0.17  & 0.17  & 0.25     & 0.39     & 0.37   & 0.43  & 0.28     & 0.22    & 0.21          & 0.20       & 0.08      & 0.17 & 0.26 & 0.30       & 0.22        & 0.27      & 0.45         & 0.25    & 0.49 & 0.27       & 0.37 & 0.23         & 0.07              & 0.74 & 0.49           & 0.50      & 0.11       \\
I-CyCLIP & RN50                    & 2.6M                  & 0.22  & 0.19  & 0.20     & 0.30     & 0.42   & 0.45  & 0.22     & 0.19    & 0.20          & 0.20       & 0.11      & 0.18 & 0.23 & 0.29       & 0.19        & 0.28      & 0.42         & 0.28    & 0.51 & 0.29       & 0.53 & 0.20         & 0.12              & 0.29 & 0.60           & 0.53      & 0.08       \\
\hline
CLIP     & RN50                    & 400M                  & 0.37  & 0.60  & 0.37     & 0.66     & 0.59   & 0.56  & 0.35     & 0.27    & 0.51          & 0.44       & 0.25      & 0.22 & 0.46 & 0.60       & 0.40        & 0.41      & 0.47         & 0.32    & 0.50 & 0.41       & 0.61 & 0.34         & 0.09              & 0.75 & 0.62           & 0.51      & 0.13       \\
CLIP     & RN101                   & 400M                  & 0.40  & 0.25  & 0.42     & 0.73     & 0.60   & 0.46  & 0.40     & 0.30    & 0.55          & 0.54       & 0.27      & 0.23 & 0.50 & 0.70       & 0.39        & 0.38      & 0.49         & 0.27    & 0.52 & 0.37       & 0.57 & 0.43         & 0.09              & 0.75 & 0.56           & 0.59      & 0.14       \\
CLIP     & RN50x4                  & 400M                  & 0.29  & 0.23  & 0.47     & 0.78     & 0.64   & 0.56  & 0.43     & 0.30    & 0.58          & 0.51       & 0.25      & 0.19 & 0.52 & 0.67       & 0.44        & 0.41      & 0.46         & 0.30    & 0.47 & 0.41       & 0.64 & 0.41         & 0.08              & 0.78 & 0.57           & 0.51      & 0.10       \\
CLIP     & RN50x16                 & 400M                  & 0.37  & 0.24  & 0.46     & 0.78     & 0.64   & 0.51  & 0.50     & 0.35    & 0.60          & 0.61       & 0.32      & 0.22 & 0.56 & 0.72       & 0.51        & 0.40      & 0.50         & 0.29    & 0.50 & 0.41       & 0.63 & 0.41         & 0.19              & 0.75 & 0.61           & 0.51      & 0.09       \\
CLIP     & RN50x64                 & 400M                  & 0.44  & 0.48  & 0.44     & 0.81     & 0.67   & 0.64  & 0.55     & 0.44    & 0.66          & 0.65       & 0.35      & 0.23 & 0.59 & 0.75       & 0.54        & 0.44      & 0.54         & 0.38    & 0.55 & 0.47       & 0.73 & 0.39         & 0.16              & 0.76 & 0.51           & 0.56      & 0.73       \\
CLIP     & ViT-B/32                & 400M                  & 0.33  & 0.26  & 0.45     & 0.79     & 0.58   & 0.49  & 0.45     & 0.41    & 0.58          & 0.55       & 0.30      & 0.25 & 0.53 & 0.71       & 0.47        & 0.42      & 0.50         & 0.35    & 0.54 & 0.40       & 0.81 & 0.50         & 0.12              & 0.75 & 0.60           & 0.53      & 0.20       \\
CLIP     & ViT-B/16                & 400M                  & 0.44  & 0.26  & 0.55     & 0.86     & 0.70   & 0.59  & 0.53     & 0.47    & 0.65          & 0.64       & 0.32      & 0.30 & 0.59 & 0.73       & 0.50        & 0.45      & 0.57         & 0.36    & 0.61 & 0.41       & 0.87 & 0.49         & 0.12              & 0.82 & 0.61           & 0.55      & 0.18       \\
CLIP     & ViT-L/14                & 400M                  & 0.52  & 0.65  & 0.52     & 0.84     & 0.63   & 0.63  & 0.59     & 0.56    & 0.68          & 0.73       & 0.37      & 0.29 & 0.62 & 0.76       & 0.51        & 0.48      & 0.56         & 0.38    & 0.53 & 0.48       & 0.68 & 0.51         & 0.11              & 0.75 & 0.62           & 0.45      & 0.30       \\
\hline
OpenCLIP & ViT-B/16                & 400M                  & 0.30  & 0.22  & 0.54     & 0.83     & 0.67   & 0.54  & 0.60     & 0.43    & 0.65          & 0.62       & 0.35      & 0.27 & 0.59 & 0.76       & 0.46        & 0.48      & 0.52         & 0.35    & 0.56 & 0.42       & 0.57 & 0.51         & 0.13              & 0.74 & 0.55           & 0.61      & 0.19       \\
OpenCLIP & ViT-B/16@240            & 400M                  & 0.41  & 0.39  & 0.57     & 0.87     & 0.65   & 0.67  & 0.62     & 0.43    & 0.65          & 0.62       & 0.34      & 0.28 & 0.61 & 0.77       & 0.49        & 0.48      & 0.56         & 0.37    & 0.58 & 0.44       & 0.57 & 0.49         & 0.09              & 0.67 & 0.57           & 0.47      & 0.18       \\
OpenCLIP & ViT-L/14                & 400M                  & 0.49  & 0.63  & 0.56     & 0.86     & 0.65   & 0.53  & 0.59     & 0.44    & 0.68          & 0.70       & 0.36      & 0.32 & 0.63 & 0.77       & 0.52        & 0.44      & 0.54         & 0.35    & 0.58 & 0.40       & 0.47 & 0.50         & 0.15              & 0.74 & 0.57           & 0.63      & 0.15       \\
OpenCLIP & ViT-L/14                & 2B                    & 0.49  & 0.41  & 0.60     & 0.90     & 0.68   & 0.60  & 0.55     & 0.47    & 0.71          & 0.73       & 0.40      & 0.32 & 0.68 & 0.81       & 0.53        & 0.49      & 0.58         & 0.39    & 0.59 & 0.47       & 0.56 & 0.54         & 0.13              & 0.74 & 0.68           & 0.67      & 0.12       \\
OpenCLIP & CoCa-ViT-L/14           & 2B                    & 0.45  & 0.61  & 0.51     & 0.90     & 0.66   & 0.63  & 0.51     & 0.61    & 0.70          & 0.73       & 0.41      & 0.33 & 0.69 & 0.81       & 0.56        & 0.48      & 0.55         & 0.38    & 0.59 & 0.45       & 0.57 & 0.51         & 0.18              & 0.74 & 0.46           & 0.53      & 0.11       \\
OpenCLIP & R-ViT-B/32        & 2B                    & 0.43  & 0.40  & 0.48     & 0.86     & 0.69   & 0.61  & 0.55     & 0.50    & 0.62          & 0.66       & 0.32      & 0.32 & 0.61 & 0.73       & 0.49        & 0.50      & 0.51         & 0.35    & 0.54 & 0.43       & 0.58 & 0.43         & 0.19              & 0.75 & 0.61           & 0.55      & 0.11       \\
OpenCLIP & XLM-R-B-ViT-B/32  & 5B                    & 0.62  & 0.57  & 0.52     & 0.87     & 0.68   & 0.58  & 0.57     & 0.50    & 0.64          & 0.60       & 0.33      & 0.30 & 0.62 & 0.73       & 0.49        & 0.46      & 0.49         & 0.35    & 0.51 & 0.43       & 0.58 & 0.47         & 0.17              & 0.75 & 0.60           & 0.62      & 0.10       \\
OpenCLIP & CoCa-ViT-B/32           & 2B                    & 0.43  & 0.53  & 0.53     & 0.84     & 0.65   & 0.59  & 0.57     & 0.40    & 0.63          & 0.64       & 0.34      & 0.29 & 0.61 & 0.73       & 0.48        & 0.45      & 0.49         & 0.36    & 0.50 & 0.40       & 0.56 & 0.45         & 0.14              & 0.75 & 0.59           & 0.62      & 0.10       \\
OpenCLIP & ViT-g/14                & 2B                    & 0.43  & 0.41  & 0.55     & 0.90     & 0.64   & 0.58  & 0.67     & 0.60    & 0.73          & 0.79       & 0.45      & 0.37 & 0.68 & 0.82       & 0.55        & 0.50      & 0.60         & 0.41    & 0.56 & 0.48       & 0.56 & 0.46         & 0.12              & 0.75 & 0.59           & 0.57      & 0.25       \\
OpenCLIP & ViT-H/14                & 2B                    & 0.43  & 0.34  & 0.61     & 0.91     & 0.66   & 0.67  & 0.63     & 0.65    & 0.75          & 0.80       & 0.45      & 0.34 & 0.70 & 0.85       & 0.57        & 0.49      & 0.60         & 0.45    & 0.63 & 0.49       & 0.47 & 0.55         & 0.22              & 0.69 & 0.68           & 0.56      & 0.12       \\
OpenCLIP & XLM-R-L-ViT-H/14  & 5.9B                    & 0.40  & 0.32  & 0.60     & 0.91     & 0.67   & 0.65  & 0.57     & 0.59    & 0.72          & 0.76       & 0.44      & 0.34 & 0.68 & 0.84       & 0.57        & 0.48      & 0.58         & 0.39    & 0.59 & 0.48       & 0.60 & 0.53         & 0.16              & 0.74 & 0.65           & 0.48      & 0.16       \\
\hline
CLIP     & ViT-B/32 FT             & 400M                  & 0.50  & 0.44  & 0.47     & 0.94     & 0.68   & 0.67  & 0.64     & 0.51    & 0.67          & 0.71       & 0.40      & 0.33 & 0.70 & 0.81       & 0.49        & 0.49      & 0.70         & 0.44    & 0.75 & 0.49       & 0.69 & 0.52         & 0.14              & 0.81 & 0.70           & 0.56      & 0.29       \\
\hline
OpenCLIP & RN50                    & 15M                   & 0.32  & 0.21  & 0.10     & 0.14     & 0.15   & 0.10  & 0.09     & 0.11    & 0.04          & 0.04       & 0.03      & 0.08 & 0.05 & 0.08       & 0.08        & 0.07      & 0.23         & 0.11    & 0.31 & 0.15       & 0.46 & 0.17         & 0.09              & 0.54 & 0.50           & 0.51      & 0.34       \\
OpenCLIP & RN50                    & 12M                   & 0.27  & 0.27  & 0.10     & 0.19     & 0.19   & 0.14  & 0.11     & 0.11    & 0.04          & 0.04       & 0.03      & 0.07 & 0.06 & 0.12       & 0.11        & 0.09      & 0.29         & 0.15    & 0.37 & 0.15       & 0.53 & 0.18         & 0.09              & 0.57 & 0.51           & 0.50      & 0.11       \\
OpenCLIP & RN101                   & 15M                   & 0.33  & 0.25  & 0.10     & 0.12     & 0.17   & 0.12  & 0.08     & 0.11    & 0.04          & 0.04       & 0.03      & 0.07 & 0.05 & 0.10       & 0.08        & 0.08      & 0.19         & 0.09    & 0.25 & 0.12       & 0.55 & 0.17         & 0.09              & 0.58 & 0.52           & 0.50      & 0.21       \\
OpenCLIP & ViT-B/32                & 400M                  & 0.30  & 0.41  & 0.41     & 0.73     & 0.62   & 0.56  & 0.52     & 0.39    & 0.56          & 0.53       & 0.28      & 0.23 & 0.53 & 0.71       & 0.41        & 0.39      & 0.51         & 0.35    & 0.54 & 0.37       & 0.39 & 0.33         & 0.05              & 0.51 & 0.52           & 0.59      & 0.45       \\
OpenCLIP & ViT-B/32                & 2B                    & 0.40  & 0.59  & 0.48     & 0.85     & 0.67   & 0.58  & 0.56     & 0.44    & 0.65          & 0.67       & 0.34      & 0.24 & 0.60 & 0.77       & 0.48        & 0.46      & 0.50         & 0.35    & 0.55 & 0.41       & 0.55 & 0.50         & 0.13              & 0.75 & 0.57           & 0.61      & 0.05       \\
OpenCLIP & CN-L@320-FT-Soup & 2B                    & 0.52  & 0.44  & 0.57     & 0.89     & 0.62   & 0.58  & 0.57     & 0.57    & 0.72          & 0.73       & 0.39      & 0.33 & 0.68 & 0.83       & 0.56        & 0.51      & 0.62         & 0.41    & 0.59 & 0.45       & 0.66 & 0.45         & 0.14              & 0.75 & 0.59           & 0.51      & 0.11       \\
OpenCLIP & CN-L@320-FT       & 2B                    & 0.53  & 0.48  & 0.56     & 0.90     & 0.62   & 0.56  & 0.55     & 0.55    & 0.71          & 0.72       & 0.37      & 0.33 & 0.67 & 0.82       & 0.55        & 0.50      & 0.63         & 0.41    & 0.60 & 0.44       & 0.66 & 0.45         & 0.14              & 0.75 & 0.58           & 0.48      & 0.10       \\
\hline
ALBEF    & ViT-B/16                & 4M                    & 0.45  & 0.37  & 0.25     & 0.41     & 0.38   & 0.43  & 0.17     & 0.22    & 0.20          & 0.27       & 0.16      & 0.22 & 0.29 & 0.29       & 0.21        & 0.36      & 0.53         & 0.30    & 0.57 & 0.34       & 0.62 & 0.21         & 0.11              & 0.74 & 0.52           & 0.51      & 0.13       \\
\hline
DeCLIP   & ViT-B/32                & 15M                   & 0.37  & 0.59  & 0.31     & 0.44     & 0.37   & 0.39  & 0.24     & 0.23    & 0.24          & 0.25       & 0.15      & 0.18 & 0.30 & 0.31       & 0.22        & 0.36      & 0.32         & 0.21    & 0.37 & 0.35       & 0.47 & 0.35         & 0.12              & 0.71 & 0.48           & 0.49      & 0.15 \\      
\hline

\end{tabular}
\end{adjustbox}
\caption{\textbf{Per-dataset results for all the remaining model configurations evaluated (see Fig. 4 in the main paper).} To reduce space in the \textit{Backbone} column, we abbreviate `Roberta' to `R' and `ConvNeXt' to `CN'.}
\label{table:remaining_results}
\end{table}

\clearpage

\subsubsection{CLIP Detailed Results}
\begin{figure*}[!ht]
  \centering
  \includegraphics[width=\linewidth]{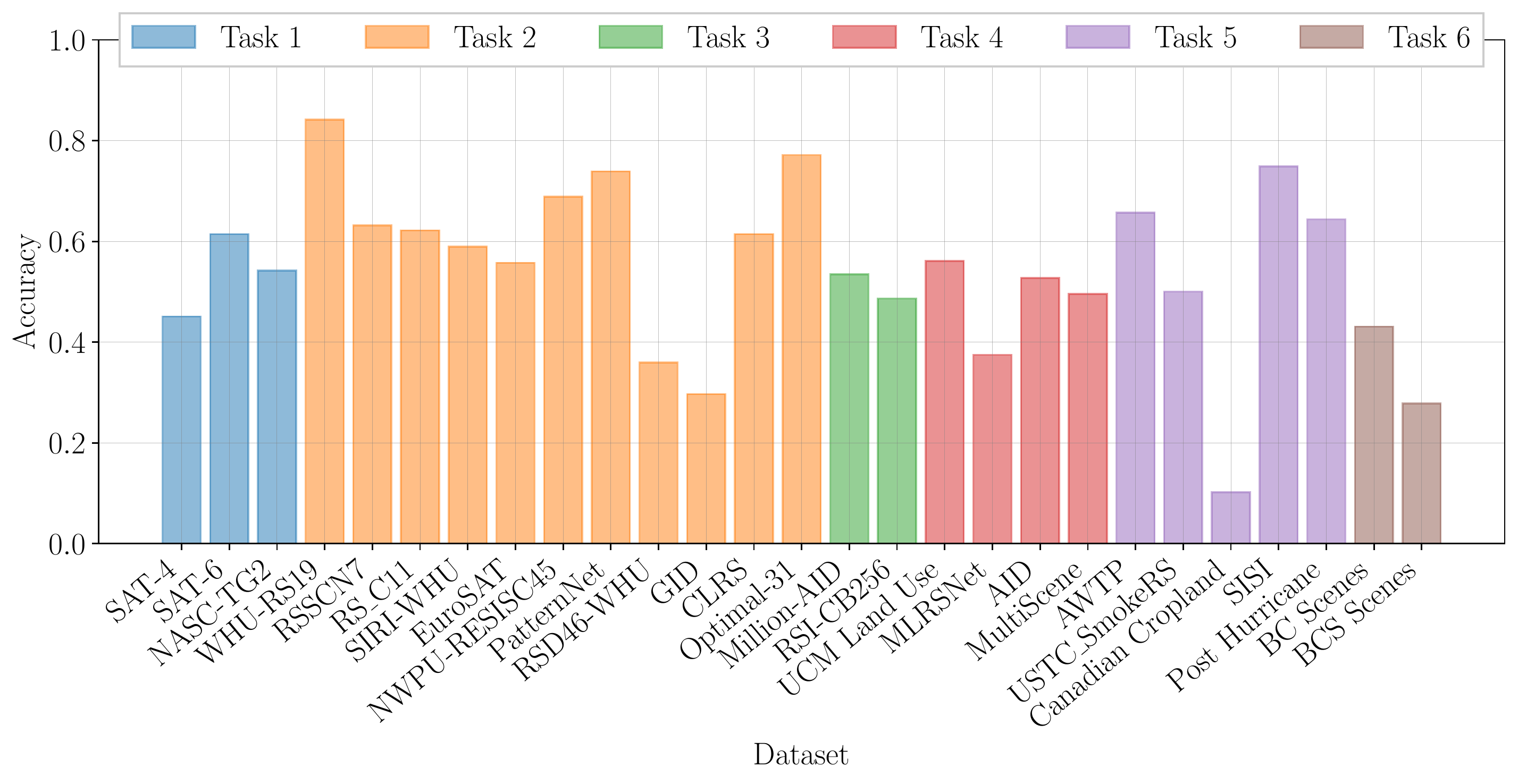}
  \vspace{-0.75cm}
  \caption{\textbf{Per-dataset accuracy scores for the CLIP 400M ViT-L/14@336px model}. Best viewed in colour.}
  \label{fig:bars}
\end{figure*}

\begin{figure*}[!ht]
\begin{subfigure}{.5\textwidth}
\includegraphics[height=6.5cm]{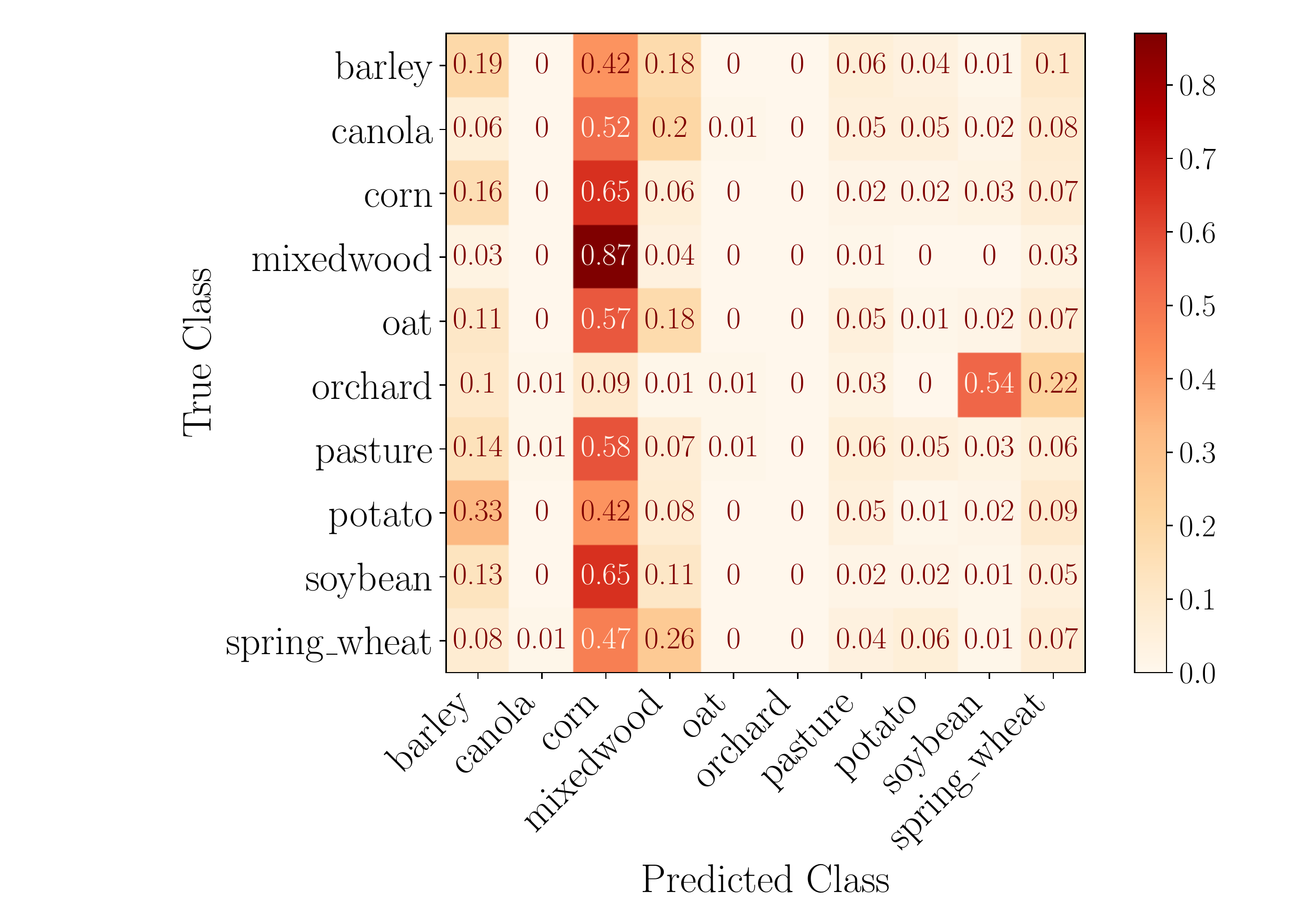}
\label{label-a}
\vspace{-0.5cm}
\caption*{\textbf{Canadian Cropland}}
\end{subfigure}
\hspace{-1.25cm}
\begin{subfigure}{.5\textwidth}
\includegraphics[height=6.5cm]{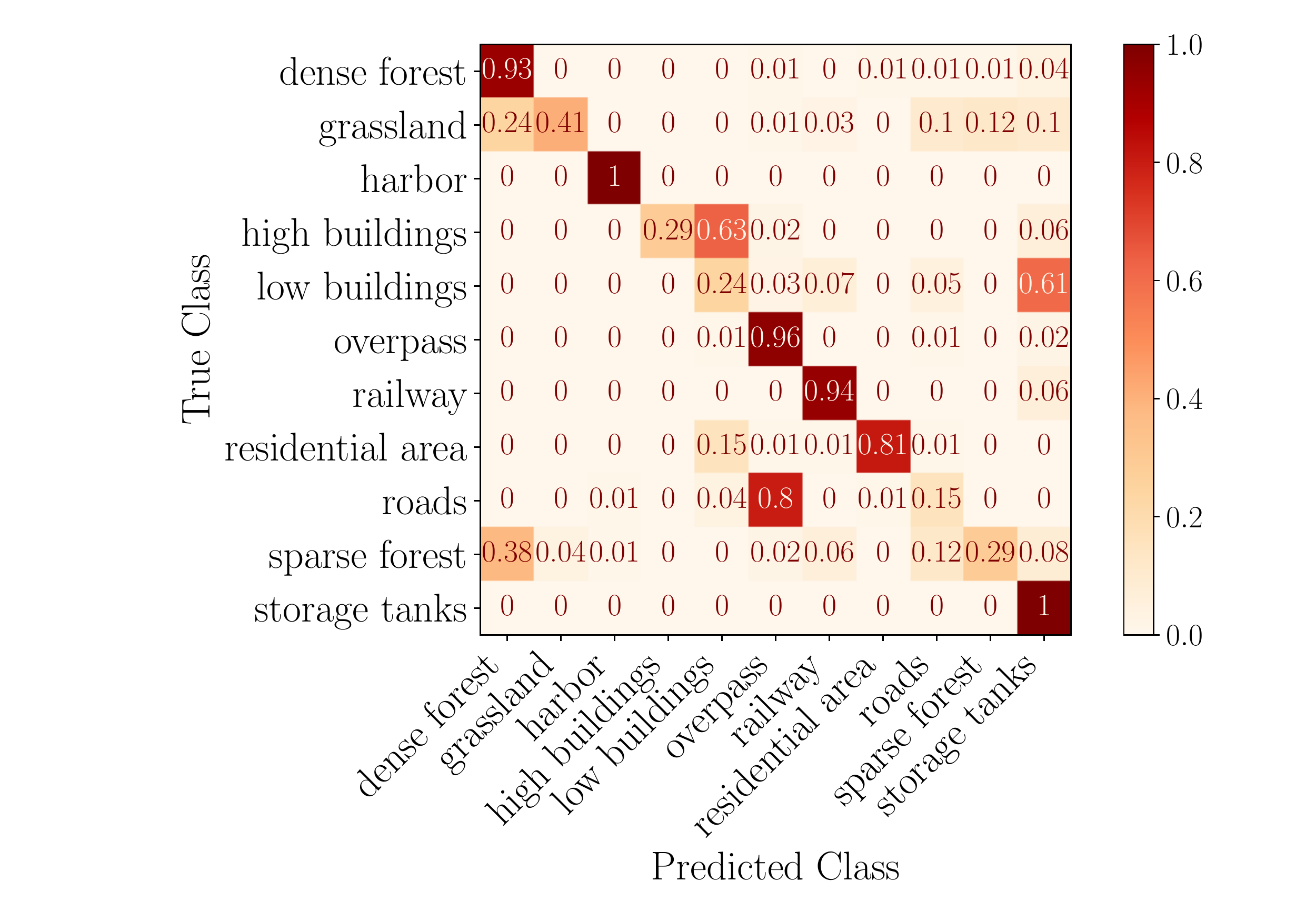}
\label{label-b}
\vspace{-0.5cm}
\caption*{\textbf{RSC11}}
\end{subfigure}
\vspace{-0.25cm}
\caption{\textbf{Confusion matrices for the CLIP ViT-L/14@336px 400M model}.}
\label{fig:clip_confusionmatrices}
\end{figure*}

Figure \ref{fig:bars} shows the accuracy attained by CLIP against the constituent SATIN datasets. The figure further illustrates the wide distribution of performances across the SATIN datasets (10\% -- 84\%) shown in Figure 3 in the main paper. There is also clear intra-task variability, and, to a lesser extent, inter-task variability. In Figure \ref{fig:clip_confusionmatrices} we display confusion matrices for a dataset that CLIP performs well on (64\% accuracy) and one that it performs poorly on (10\% accuracy). The Canadian Cropland dataset is especially challenging because it contains different classes (crop species) that are very closely related and difficult to distinguish. The confusion matrix reflects this difficulty as almost all predictions are made against a single class, \textit{corn}. The RSC11 dataset includes more diverse land use categories, which, in most cases, CLIP is able to tell apart. However, it can be seen to struggle on closely related classes such as \textit{dense forest} and \textit{spare forest}, as well as, \textit{overpass} and \textit{roads}.


\vspace{-0.25cm}

\clearpage

\subsection{Methodology}
\label{section:methodology}

This section provides a more detailed description of the methodology we took to attain the results we presented in the main paper. Specifically, we provide PyTorch- and HuggingFace-style pseudocode for our inference pipeline in Pseudocode
\ref{pseudocode:inference}.
To enable complete reproducibility, we also include links to the model checkpoints and implementation scripts that we used for each of the models we evaluated (see Table \ref{table:model_checkpoints}) as well as the exact prompt templates used per task (see Table \ref{table:prompt_templates}).

\subsubsection{Inference Pseudocode}
\label{appendix:pseudocode}

\begin{algorithm}[!h]
\SetAlgoLined
    \PyComment{dataset\_name \,\,\enskip\quad\quad - SATIN dataset (e.g., EuroSAT)} \\
    \PyComment{prompt\_template \quad - Textual prompt string (e.g., "A satellite photo of ")} \\
    \PyComment{k \,\qquad\qquad\qquad\qquad\quad - Number of true labels for each image} \\
    \PyComment{model \,\enskip\qquad\qquad\qquad - Pretrained VL model} \\
    \PyComment{processor \,\,\qquad\qquad - Image/text pre-processor} \\
    \PyCode{} \\
    \PyComment{0. Load data} \\
    \PyCode{dataset = load\_dataset(path="SATIN", split=dataset\_name)} \PyComment{download dataset} \\
    \PyCode{images = dataset.features["image"]} \PyComment{extract images from dataset} \\
    \PyCode{labels = dataset.features["label"]} \PyComment{extract labels from dataset} \\
    \PyCode{class\_names = dataset.features["label"].names} \PyComment{extract class names} \\
    \PyCode{} \\
    \PyComment{1. Preprocess input images and text} \\
    \PyCode{text\_prompts = [prompt\_template + c for c in class\_names]} \PyComment{convert to prompts} \\
    \PyCode{text\_inputs = processor(text=text\_prompts)} \\
    \PyCode{image\_inputs = processor(images=images)} \\
    \PyCode{} \\
    \PyComment{2. Extract image and text features} \\
    \PyCode{text\_features = model.get\_text\_features(**text\_inputs)} \\
    \PyCode{image\_features = model.get\_image\_features(**image\_inputs)} \\
    \PyCode{} \\
    \PyComment{3. Normalise features} \\
    \PyCode{text\_features /= text\_features.norm(dim=-1, keepdim=True)} \\
    \PyCode{image\_features /= image\_features.norm(dim=-1, keepdim=True)} \\
    \PyCode{} \\
    \PyComment{4. Compute similarity} \\
    \PyCode{similarity = (100.0 * image\_features @ text\_features.T).softmax(dim=-1)} \\
    \PyCode{\_, indices = similarity.topk(k)} \\ 
    \PyCode{} \\
    \PyComment{5. Determine top predicted class(es)} \\    
    \PyComment{Correct predictions if indices == label} \\    
    
\vspace{0.5cm}
\SetAlgorithmName{Pseudocode}
\SetAlgoRefName{Pseudocode}
\caption{\textbf{PyTorch- and HuggingFace-style pseudocode outlining the general procedure for inference on SATIN}. For particular models or implementations, steps 1 and 2 vary slightly in syntax but the functionality taken at each step remains the same. The comment numbers refer to the inference steps outlined in the main paper.}
\label{pseudocode:inference}
\end{algorithm}

\clearpage

\subsubsection{Model Checkpoints}

\begin{table}[!ht]
    \small
    \centering
    \begin{tabular}{lll|p{7.5cm}}
    \hline
    \textbf{Method} & \textbf{Backbone} & \textbf{Pretraining} & \textbf{Checkpoint source}\\
    \hline
    CyCLIP, CLIP, I/C-CyCLIP & RN50 & 500K -- 3M & \url{https://github.com/goel-shashank/CyCLIP} \\   
    \hline
    CLIP & All remaining & 400M & \url{https://github.com/openai/CLIP} \\
    \hline
    OpenCLIP & ConvNeXt-based & 12M -- 5B & \url{https://huggingface.co/laion} \\
    \hline
    OpenCLIP & All remaining & 12M -- 5B & \url{https://github.com/mlfoundations/open_clip} \\
    \hline
    ALBEF & ViT-B/16 & 4M & \url{https://github.com/salesforce/ALBEF} \\
    \hline
    ALBEF & ViT-B/16 & 14M & \url{https://github.com/salesforce/LAVIS} \\
    \hline
    BLIP & ViT-B/32 & 129M & \url{https://github.com/salesforce/LAVIS} \\
    \hline
    BLIP2 & ViT-G/14 & 129M & \url{https://github.com/salesforce/LAVIS} \\
    \hline
    DeCLIP & ViT-B/32 & 15M, 88M & \url{https://github.com/Computer-Vision-in-the-Wild/Elevater_Toolkit_IC} \\
    \hline
    SLIP & ViT-B/32 & 15M & \url{https://github.com/Computer-Vision-in-the-Wild/Elevater_Toolkit_IC} \\
    \hline
    CLIP Fine-tuned & ViT-B/32 & 400M + 13K & \url{https://huggingface.co/flax-community/clip-rsicd-v2} \\  
    \hline
    \end{tabular}
    \vspace{0.25cm}
    \caption{\textbf{Model checkpoints}. We provide a link for checkpoint and implementation scripts for each of the models we evaluated.
    }
    \label{table:model_checkpoints}
\end{table}

\subsubsection{Prompt Templates}
\begin{table}[!ht]
    \small
    \centering
    \begin{tabular}{l|l}
    \hline
    \textbf{Task} & \textbf{Prompts}\\
    \hline
    \multirow{2}{*}{Task 1. \textbf{Land Cover}} & ``a satellite photo of \{\textit{CLS}\}"\\ & ``an aerial photo of \{\textit{CLS}\}" \\   
    \hline
    \multirow{2}{*}{Task 2. \textbf{Land Use}} & ``a satellite photo of \{\textit{CLS}\}"\\ & ``an aerial photo of \{\textit{CLS}\}" \\  
    \hline
    \multirow{2}{*}{Task 3. \textbf{Hierarchical Land Use}} & ``a satellite photo of \{\textit{CLS}\}"\\ & ``an aerial photo of \{\textit{CLS}\}" \\ 
    \hline
    \multirow{2}{*}{Task 4. \textbf{Complex Scenes}} & ``a satellite photo containing \{\textit{CLS}\}"\\ & ``an aerial photo containing \{\textit{CLS}\}" \\ 
    \hline
    \multirow{2}{*}{Task 5. \textbf{Rare Scenes}} & ``a satellite photo of \{\textit{CLS}\}"\\ & ``an aerial photo of \{\textit{CLS}\}" \\ 
    \hline
    \multirow{2}{*}{Task 6. \textbf{False-colour Scenes}} & ``a false-colour satellite photo of \{\textit{CLS}\}"\\ & ``a false-colour aerial photo of \{\textit{CLS}\}" \\ 
    \hline    
    \end{tabular}
    \vspace{0.25cm}
    \caption{\textbf{Prompt templates}. Inference was carried out on all the datasets in a given task using the same prompt templates.
    }
    \label{table:prompt_templates}
\end{table}

\clearpage

\subsection{Compute Requirements}
\label{appendix:compute}
\label{section:compute_requirements}
For all of our experiments, we carried out inference on a single Nvidia A100-80GB GPU with a batch size of 128. For each of the models evaluated in Table 3 of the main paper, we benchmark the inference time and maximum GPU memory required when evaluating across the entire SATIN benchmark (see Table \ref{table:compute_requirements}). Additionally, we further provide the per-dataset inference times taken for each model (see Table \ref{table:compute_requirements_per_dataset}). Note, as with the accuracy results outlined in this supplementary material and in the main paper, different model configurations are used for the different methods, therefore this analysis does not offer a direct comparison between the relative compute requirements of the different methods.

\begin{table}[!ht]
    \small
    \centering
    \begin{tabular}{ll|cc}
    \hline
    \textbf{Method} & \textbf{Backbone} & \textbf{Inference Time (minutes)} & \textbf{GPU Memory (GB)} \\
    \hline
    CyCLIP & RN50 & 30.6 & 5.02 \\
    ALBEF & ViT-B/16 & 25.6 & 5.14 \\  
    SLIP & ViT-B/32 & 17.2 & 3.63 \\  
    DeCLIP & ViT-B/32 & 18.5 & 3.30 \\  
    BLIP & ViT-B/16 & 26.8 & 4.97 \\  
    BLIP2 & ViT-G/14 & 106.3 & 11.79 \\  
    CLIP & ViT-L/14@336px & 127.0 & 8.41 \\  
    OpenCLIP & ViT-G/14 & 155.4 & 17.42 \\  
    
    \hline
    \end{tabular}
    \vspace{0.25cm}
    \caption{\textbf{Compute requirements of inference on SATIN for each of the methods evaluated in this work}.
    }
    \label{table:compute_requirements}
\end{table}

\begin{table}[!ht]
    \small
    \centering
    \begin{tabular}{ l | c c c c c c c c }
    \hline
    \multicolumn{1}{c}{\textbf{Dataset}} & \multicolumn{8}{c}{\textbf{Inference Time (seconds)}} \\
    \hline
    \textbf{} & \textbf{CyCLIP} & \textbf{ALBEF} & \textbf{SLIP} & \textbf{DeCLIP} & \textbf{BLIP} & \textbf{BLIP2} & \textbf{CLIP} & \textbf{OpenCLIP} \\
    \hline
SAT-4 & 118.8 & 143.0 & 59.7 & 60.0 & 144.9 & 882.5 & 1073.4 & 1389.8 \\
SAT-6 & 97.4 & 115.4 & 48.5 & 52.1 & 121.0 & 720.8 & 873.9 & 1158.4 \\
NaSC-TG2 & 35.5 & 38.5 & 21.0 & 26.3 & 43.4 & 198.9 & 233.8 & 276.6 \\
\hline
WHU-RS19 & 12.4 & 12.2 & 12.3 & 17.5 & 18.0 & 27.9 & 33.6 & 24.0 \\
RSSCN7 & 28.7 & 19.2 & 17.1 & 22.6 & 23.7 & 47.0 & 54.1 & 52.1 \\
RSC11 & 35.3 & 10.6 & 10.9 & 14.6 & 14.8 & 27.7 & 31.1 & 25.5 \\
SIRI-WHU & 14.6 & 14.3 & 13.3 & 18.2 & 19.3 & 40.8 & 45.6 & 43.2 \\
EuroSAT & 39.9 & 45.6 & 22.9 & 28.6 & 51.1 & 252.3 & 304.8 & 368.7 \\
NWPU-RESISC45 & 79.6 & 74.5 & 48.7 & 50.8 & 77.0 & 313.2 & 376.4 & 456.7 \\
PatternNet & 125.0 & 98.9 & 72.6 & 74.1 & 96.0 & 327.5 & 386.8 & 471.4 \\
RSD46-WHU & 151.9 & 98.2 & 86.3 & 83.2 & 97.0 & 229.1 & 265.5 & 313.9 \\
GID & 101.1 & 69.3 & 45.0 & 48.7 & 70.6 & 294.3 & 355.6 & 439.9 \\
CLRS & 131.2 & 48.8 & 36.9 & 38.4 & 47.4 & 166.9 & 195.1 & 228.9 \\
Optimal-31 & 17.8 & 12.8 & 12.1 & 16.7 & 15.9 & 31.8 & 37.3 & 35.4 \\
\hline
RSI-CB256 & 132.1 & 72.7 & 52.3 & 52.4 & 70.6 & 258.8 & 307.2 & 365.7 \\
Million-AID & 75.1 & 65.8 & 59.3 & 64.1 & 71.4 & 147.4 & 163.8 & 186.7 \\
\hline
UCM Land Use & 21.3 & 16.8 & 16.0 & 16.8 & 20.0 & 36.9 & 40.0 & 42.9 \\
MLRSNet & 228.0 & 239.7 & 149.7 & 146.6 & 243.8 & 1052.6 & 1262.3 & 1561.6 \\
AID & 30.0 & 31.2 & 30.6 & 33.3 & 33.4 & 59.4 & 63.3 & 71.0 \\
MultiScene & 94.0 & 69.3 & 61.0 & 60.9 & 71.0 & 175.9 & 204.9 & 241.7 \\
\hline
AWTP & 104.9 & 121.9 & 62.5 & 64.5 & 121.0 & 653.2 & 789.2 & 980.7 \\
USTC-SmokeRS & 54.1 & 24.8 & 20.4 & 24.6 & 25.9 & 75.6 & 89.8 & 98.7 \\
Canadian Cropland & 31.4 & 30.5 & 20.1 & 24.6 & 33.4 & 139.9 & 169.0 & 204.0 \\
SISI & 22.5 & 15.7 & 13.2 & 17.9 & 18.4 & 50.4 & 60.2 & 65.3 \\
Post Hurricane & 24.8 & 24.6 & 17.5 & 21.7 & 28.3 & 103.2 & 124.8 & 144.8 \\
\hline
BC Scenes & 15.6 & 13.1 & 11.7 & 16.8 & 15.7 & 39.7 & 46.1 & 47.0 \\
BCS Scenes & 13.2 & 10.8 & 11.3 & 15.5 & 14.0 & 26.2 & 30.1 & 26.9 \\
\hline
\hline
    \end{tabular}
    \vspace{0.25cm}
    \caption{\textbf{Per-dataset inference times for the models outlined in Table \ref{table:compute_requirements}}. 
    }
    \label{table:compute_requirements_per_dataset}
\end{table}

\end{document}